\documentclass[sensors,article,accept,pdftex,moreauthors]{Definitions/mdpi} 

\firstpage{1} 
\pubvolume{1}
\issuenum{1}
\articlenumber{0}
\pubyear{2022}
\copyrightyear{2022}
\externaleditor{Academic Editor: Gregor Klancar, Marija Seder, Sašo Blažič 
}
\datereceived{17 July 2022} 
\dateaccepted{25 August 2022} 
\datepublished{} 
\hreflink{https://doi.org/} 

\usepackage[labelformat=simple]{subcaption}

\DeclareCaptionLabelFormat{subcaptionlabel}{\normalfont(\textbf{#2}\normalfont)}
\usepackage[flushleft]{threeparttable}
\usepackage{multirow}

\usepackage[normalem]{ulem}
\renewcommand{\sout}{\bgroup \ULdepth=-.45ex \ULset} 
\newcommand\suppress[1]{} 
\newcommand\myOptElse[3]{\ifthenelse{\boolean{#1}}%
{#2\suppress{#3}}%
{\suppress{#2}#3}}

\definecolor{commentcolor}{gray}{0.5}
\usepackage{algorithm}
\usepackage{algpseudocode}
\algrenewcommand\algorithmicindent{1.0em}%

\algnewcommand{\LineComment}[1]{\State \textcolor{commentcolor}{\(\triangleright\) #1}}
\algnewcommand{\To}{\textbf{to}}
\algnewcommand{\Break}{\textbf{break}}
\algnewcommand{\Continue}{\textbf{continue}}
\algnewcommand{\IIf}[1]{\State\algorithmicif\ #1\ \algorithmicthen}
\algnewcommand{\EndIIf}{\unskip}
\algnewcommand{\var}[1]{\textit{#1}}
\algnewcommand{\func}[1]{\textsc{#1}}

\Title{Visual Servoing Approach to Autonomous UAV Landing on a Moving Vehicle}

\TitleCitation{Visual Servoing Approach to Autonomous UAV Landing on a Moving Vehicle}

\Author{Azarakhsh Keipour 
$^{1,*,\dagger}$\orcidA{}, Guilherme A. S. Pereira 
 $^{2}$\orcidB{}, Rogerio Bonatti $^{3}$\orcidC{}, Rohit Garg $^{4}$, Puru Rastogi $^{5}$, \linebreak Geetesh Dubey $^{4}$, and Sebastian Scherer $^{6}$\orcidD{}}

\AuthorNames{Azarakhsh Keipour, Guilherme A. S. Pereira, Rogerio Bonatti, Rohit Garg, Puru Rastogi, Geetesh Dubey and Sebastian Scherer}

\AuthorCitation{Keipour, A.; 
Pereira, G.A.S.; Bonatti, R.; Garg, R.; Rastogi, P.; Dubey, G.; Scherer, S.}

\address{%
$^{1}$ \quad Robotics AI, 
Amazon, Arlington, VA 22202, USA\\
$^{2}$ \quad Department of Mechanical and Aerospace Engineering, West Virginia University, Morgantown,~WV~26506,~USA; guilherme.pereira@mail.wvu.edu\\
$^{3}$ \quad Autonomous Systems Group, Microsoft, Redmond, WA 98052, USA; rbonatti@microsoft.com\\
$^{4}$ \quad Gather AI, Pittsburgh, PA 15217, USA; 
rohitgarg617@gmail.com (R.G.); geeteshdubey@gmail.com (G.D.)\\
$^{5}$ \quad Mowito, 
Bengaluru 560091, 
India; pururastogi@gmail.com\\
$^{6}$ \quad Robotics Institute, Carnegie Mellon University, Pittsburgh, PA 15213, USA; basti@cmu.edu}

\corres{Correspondence: keipour@gmail.com; Tel.: +1-412-273-2712}

\firstnote{The publication 
was written prior to A. Keipour joining Amazon. During the realization of this work, he was affiliated with the Robotics Institute at Carnegie Mellon University.} 

\abstract{Many aerial robotic applications require the ability to land on moving platforms, such as delivery trucks and marine research boats. We present a method to autonomously land an Unmanned Aerial Vehicle on a moving vehicle. A visual servoing controller approaches the ground vehicle using velocity commands calculated directly in image space. The control laws generate velocity commands in all three dimensions, eliminating the need for a separate height controller. The method has shown the ability to approach and land on the moving deck in simulation, indoor and outdoor environments, and compared to the other available methods, it has provided the fastest landing approach. Unlike many existing methods for landing on fast-moving platforms, this method does not rely on additional external setups, such as RTK, motion capture system, ground station, offboard processing, or communication with the vehicle, and it requires only the minimal set of hardware and localization sensors. The videos and source codes are also provided.}


\begin{document}


\section{Introduction} \label{sec:introduction}

The recent advances in Unmanned Aerial Vehicles (UAVs) have allowed innovative applications ranging from package delivery to infrastructure inspection, early fire detection, and cinematography~\cite{bonatti, Keipour:2020:arxiv:integration, Keipour:2022:thesis}. While many of these applications require landing on the ground and static platforms, the ability to land on dynamic platforms is essential for some other applications. A typical example of a real-world scenario is an autonomous landing on a ship deck~\cite{sankalp2013} or maritime Search and Rescue operations~\cite{Almeshal2018}.

The autonomous landing of Unmanned Aerial Vehicles (UAVs) on known patterns has been an active area of research for several years~\cite{Saripalli2006, wenzel2011, alvika2014, Kim2014, bahnemann2018, beul2018, Alarcon2019}. Some of the key challenges of the problem include dealing with environmental conditions, such as changes in light and wind, and robust detection of the landing zone. The subsequent maneuver in trying to land also needs to take care of the potential ground effects at the proximity of the landing surface.

\begin{figure}[H]
\includegraphics[width=0.3\linewidth]{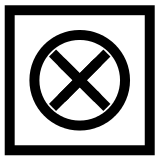}
\caption{Landing zone pattern utilized at the MBZIRC competition~\cite{mbzirc2015}.}
\label{fig1}
\end{figure}

The Mohamed Bin Zayed International Robotics Challenge (MBZIRC) is a set of real-world robotics challenges happening every few years~\cite{bhattacharya2021}. Challenge~1 took place in March 2017, focusing on landing UAVs on a moving platform. In this challenge, there was a ground vehicle (truck) moving on an 8-shaped road in a $90\times60$~m arena with a predefined speed of $15$~km/h ($4.17$~m/s). On top of the truck, at $1.5$~m height, there was a flat horizontal ferromagnetic deck with a predefined $1.5\times1.5$~m pattern (Figure~\ref{fig1}) printed on top of it~\cite{mbzirc2015}. The goal was to land the UAV on this deck autonomously. To make the challenge realistic, no communication between the UAV and the ground vehicle was allowed, no precise state estimation sensors such as RTK or Motion Capture were provided, and finally, the location of the vehicle was not given to the UAV, requiring a visual detection of the pattern, which could result in false and imperfect detections among other issues.

This paper describes our approach to landing on moving platforms, which uses only a single monocular camera, a point Lidar, and an onboard computer. The method is based on a visual-servoing controller and has been successfully tested for landing on moving ground vehicles at speeds of up to $15$~km/h ($4.17$~m/s).

Our contributions include proposing a method that does not depend on a special environment setup (e.g., IR markers and communication channels), can work with the minimal UAV sensor setup, only requires a monocular camera, has minimum processing requirements (e.g., a lightweight onboard computer), and approaches the landing platform with high certainty. We also provide our source codes and simulation environment to assist with future developments. The videos of our experiments and all the codes are available at \url{http://theairlab.org/landing-on-vehicle}. 

The paper is organized as follows: 
Section~\ref{sec:background} reviews the relevant work and state of the art; Section~\ref{sec:approach} explains our method for autonomous landing; Section~\ref{sec:results} describes the architecture and the experiments performed and discusses the results from simulation, indoor and outdoor tests.


\section{Background and Relevant Research} \label{sec:background}

A review of the existing research reveals several approaches to solving this problem. 
One of the earliest vision-based approaches introduced an image moment-based method to land a scale helicopter on a static landing pad that has a distinct geometric shape~\cite{Saripalli2003}. This work further extends to the helicopter landing on a moving target~\cite{Saripalli2006}. However, a limitation of the approach is that it tracks the target in a single dimension, and the image data are processed offboard.

A more recent work tries to land a quadcopter on top of a ground vehicle traveling at high speed using an AprilTag fiducial marker for landing pad detection~\cite{leny2017}. They devise a combination of Proportional Navigation (PN) guidance and Proportional-Derivative (PD) control to execute the task. The UAV has landed on a target moving at up to $50$~km/h ($13.89$~m/s). However, the PN guidance strategy used to approach the target has to rely on the wireless transmission of GPS and IMU measurements from the ground vehicle to the UAV, which is impractical in most applications. In addition,  two cameras are used for the final approach to detect the AprilTag from far and near. 

Visual servoing control is another approach that can be applied to this problem. In~\cite{dlee2012}, it is used to generate a velocity reference command to the lower level controller of a quadrotor, guiding the vehicle to a target landing pad moving at $0.07$~m/s. It is computationally cheaper to compute the control signals in image space than in 3D space. However, this work relies on offboard computing as well as a VICON motion capture system for accurate position feedback during its patrol to search for the target. Another closed-loop approach for landing was proposed in~\cite{9093162}. In this paper, the authors presented a landing vector field. Unlike visual servoing, the method's main advantage is that the vector field can enforce the shape of the vehicle's trajectory during landing. In addition, different from visual servoing, the 3D localization of the UAV is necessary. 

Some other works emphasize the design of a landing pad that is robust to detection, which simplifies the task by eliminating the detection uncertainty~\cite{lange2009, conte2006, Xing2019}. In~\cite{wenzel2011}, a Wii (IR) camera is used to track a T-shaped 3D pattern of infrared lights mounted on the vehicle, and, in~\cite{9189835}, an IR beacon is placed on the landing pad. Reliable detection enables a ground-based system to estimate the position of the UAV relative to the landing target. Xing et al.~\cite{Xing2019} propose a notched ring with a square landmark inside to enable robust detection of the landing zone, then it uses a minimum-jerk trajectory to land on the detected pattern. Finally, the method presented in~\cite{wynn2019visual} localizes the vehicle using two nested ArUco markers~\cite{Ramirez18} in an illuminated landing pad, allowing landing during the night using visual servoing.

Optical flow is another technique used to provide visual feedback for guiding the UAV. Ruffier and Franceschini~\cite{ruffier2015} developed an autopilot that uses a ventral optic flow regulator in one of its feedback loops for controlling lift (or altitude). The UAV can land on a platform that moves along two axes. The method in~\cite{herisse2012} also uses the optical flow information obtained from a textured landing target but does not attempt to reconstruct velocity or distance to the goal. Instead, the approach chooses to control the UAV within the image-based paradigm. 

The method in~\cite{s22010404} transfers all the computing power to the ground station eliminating the need for an onboard computer. They only report results for very slow vehicle speeds (10--15~cm/s), and the method cannot extend to higher speeds due to communication~delays.

Several approaches were introduced for the MBZIRC challenge. The method used by~\cite{bahnemann2018} tries to follow the platform until the pattern detection rate is high enough for landing. Then, it continues to follow the platform on the horizontal plane while slowly decreasing the altitude of the UAV. A Lidar (Light Detection and Ranging sensor) is used to determine if the UAV should land on the platform or not. The landing is performed by a fast descent followed by switching off the motors. The state estimation uses monocular VI-Sensor data fused with GPS, IMU, and RTK data.

Researchers from the University of Catania implemented a system that detects and tracks the target pattern using a Tracking-Learning-Detection-based method integrated with the Circle Hough Transform to find the precise location of the landing zone. Then, a Kalman filter is used to estimate the vehicle's trajectory for the UAV to follow and approach it for landing~\cite{Cantelli2017, Battiato2017}.

Baca et al.~\cite{Baca2019} took advantage of a SuperFisheye high-resolution monocular camera with a high FPS rate for pattern detection. They used adaptive thresholding to improve the method's robustness to the light intensity, followed by undistorting the image. Then, the circle and the cross are detected to find the landing pattern in the frame. After applying a Kalman filter on the detected coordinates to estimate and predict the location of the ground vehicle, a model-predictive controller (MPC) is devised to generate the reference trajectory for the UAV in real-time, which is tracked by a nonlinear feedback controller to approach and land on the ground vehicle. 

Researchers at the University of Bonn have achieved the fastest approach for landing among the other successful runs in the MBZIRC competition (measured from the time of detection to the successful landing). They used two cameras for the landing pattern detection, a Nonlinear Model Predictive Control (MPC) for time-optimal trajectory generation and control, and a separate proportional controller for yaw \cite{beul2018}.

Tzoumanikas et al.~\cite{tzoumanikas2018} used an RGB-D camera for visual-inertial state estimation. After the initial detection, the UAV flies in the vehicle's direction until it reaches the velocity and position close to the ground vehicle. Then, the UAV starts descending toward the target until a certain altitude, when it will continue descending in an open-loop manner. During the landing, the UAV uses pattern detection feedback only to decide between aborting the mission or not and not for correcting controller commands. 

University of Zurich researchers have introduced a system in~\cite{Falanga2017} that first finds the quadrangle of the pattern and then searches for the ellipse or the cross, validating the detection using RANSAC. Then, the platform's position is estimated from its relative position to the UAV, and an optimal landing trajectory is generated to land the UAV on the~vehicle.

An ultimate goal in robotics' real-world applications is to achieve good results while reducing the robot's costs and the amount of the prior setup. Dependence on the additional hardware or external hardware (e.g., motion capture systems, RTK, multiple cameras) is costly and not always feasible in real-world applications. In contrast with other available methods, our approach uses only a single monocular camera (with comparatively low resolution and low frame rate), a point Lidar, and an onboard computer to land on the moving ground vehicle at high speeds (tested at $15$~km/h ($4.17$~m/s)) using a visual-servoing controller. The approach does not depend on accurate localization sensors (e.g., RTK or motion capture systems) and can work with the state estimation provided by a commercial UAV platform (obtained only using GPS, IMU, and barometer data). Our real-time elliptic pattern detection and tracking method can track the landing deck in challenging environment conditions (e.g., changes in lighting) with a high frequency~\cite{Keipour:2021:ral:ellipse}.

\section{Materials and Methods} \label{sec:approach}

This section explains our approach to sensing the vehicle passing below the UAV and then landing the quadrotor on the vehicle. Our system was developed in the MBZIRC Challenge~1 context, discussed in Section~\ref{sec:introduction}. Therefore, our primary goal was to land on a known platform (see Figure~\ref{fig1}) fixed on top of a vehicle moving at a fixed speed.

\subsection{General Strategy} \label{sec:landingstrategy}

In our approach, we assume that the drone is hovering above the point where the moving vehicle is expected to pass (which can be any point on the road if the vehicle is in a loop). In addition,  we assume that the direction and the speed of the moving vehicle are known, which can be previously obtained, for example, from some consecutive deck pattern detections. 
We further assume that the vehicle is almost moving in a straight line during the few seconds when the UAV attempts to land. 

A time-optimal trajectory to landing is theoretically possible, while in practice, there is a variation in the speed of the moving vehicle (which is driven by a human driver), and there are delays and errors in state estimation. Therefore, an approach with constant feedback to correct the trajectory would work better than an open-loop landing using an optimal trajectory. The method we chose is to continually measure the target's position, size, and orientation directly in the image space and translate it to velocity commands for the UAV using visual servoing. This approach is described in Section~\ref{sec:landingvisualservo}.

Given the assumptions mentioned above, the defined strategy is as follows:

\begin{enumerate}
\item The quadrotor hovers at the point at the height of $h$ meters until it senses the passing ground vehicle. 

\item As soon as the ground vehicle is sensed, the UAV starts flying in the direction of movement with a predefined speed slightly higher than the vehicle speed to compensate for the distance between the UAV and the vehicle. This gap results from the delays in processing the sensors and dynamics of the UAV in getting up to the vehicle speed. 

\item After the initial acceleration, the UAV flies with a feed-forwarding speed equal to the ground vehicle speed. Then, the visual servoing controller tries to decrease the remaining gap between the UAV and the target. 

\item After successfully landing, magnets at the bottom of the UAV legs stick to the metal platform, and the propellers are shut down. 
\end{enumerate}

\begin{figure}[H]
\includegraphics[width=\linewidth]{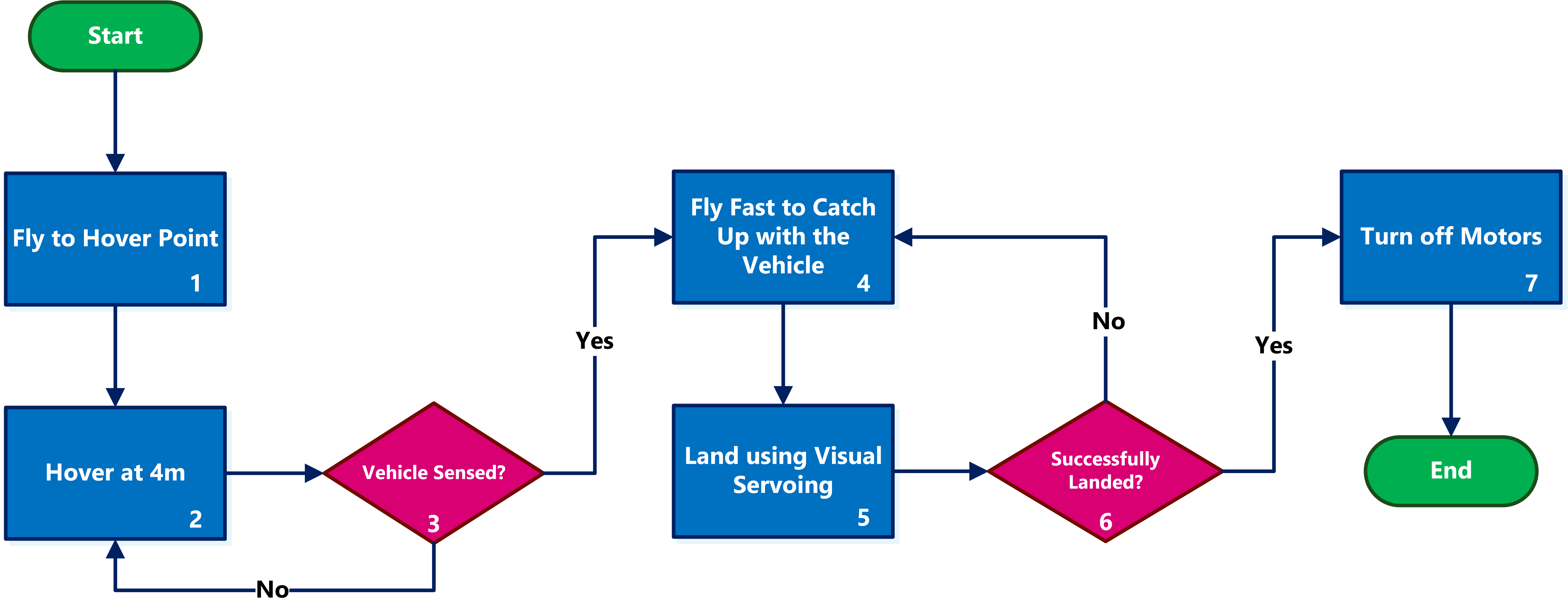}
\caption{Our strategy for landing on a moving vehicle. The vehicle hovers above the road until it senses the passing vehicle. Then, it accelerates to catch up with the vehicle and tries to land using the visual servoing method. As soon as the landing is detected, all the motors are turned off. }
\label{fig2}
\end{figure}

Figure~\ref{fig2} depicts the steps of the described strategy. The following subsections explain the main components of the landing strategy in greater detail.

We used a path tracker similar to the one proposed by \cite{hoffmann2008quadrotor} to fly to the hover point and to fly the straight lines.

\subsection{Sensing the Passing Vehicle} \label{sec:trigger}

We considered two strategies for detecting the passing vehicle below the UAV when it is hovering above the road on the vehicle's path: the camera or using laser sensors. Due to the reliability of our pattern detection method, using the camera gives more reliable results. The chosen hovering altitude can range from a few to tens of meters, where the pattern detection is reliable. The lower altitude allows for a faster approach to landing, while the higher altitude results in more flight time required to catch up with the vehicle (box 4 in Figure~\ref{fig2}). 

The strategy of sensing the vehicle using a camera worked well in our tests; however, for higher vehicle speeds, we developed a laser-based solution to avoid the delays introduced by the image processing and to get an even lower reaction time. We used three point-laser sensors on the bottom of the UAV (as described in Section~\ref{sec:hardware}) to detect the vehicle passing below it. If the distance measured by any of the three lasers has a sudden drop of more than a certain threshold within a specified period, we assume that the vehicle is passing below the quadrotor, and the landing system is triggered.

The choice of the number of the lasers and the ideal height of the hovering depend on the road's width and the vehicle's width and height. For our tests, the road was $3$~m wide, the vehicle was $1.5$~m wide, and its height was $1.5$~m. In order to capture this vehicle, the center laser points directly downwards, and the other two lasers are angled at approximately $30$~degrees on each side. With this configuration, at the altitude of $4$~m, two additional laser rays will hit the ground at a distance of $1.44$~m from the center laser. When the moving vehicle (with a width and height of $1.5$~m) passes on the road below the quadrotor, at least one of the three lasers will sense the change in the measured height. Algorithm~\ref{alg1} illustrates the method.

\begin{algorithm}[H]
\caption{Approach 
 for sensing the passing vehicle using three point lasers.}
\label{alg1}
\begin{algorithmic}[1]
\Function{DetectPassingVehicle}{}
    \LineComment {Keep reading the sensors until the vehicle is sensed}
	\While {\textbf{true}}
	    \LineComment {Read the laser measurements}
        \State $dist_l$, $dist_c$, $dist_r$ $\gets$ \textsc{ReadLasers}()
        
    	\LineComment {Check if we have a sudden decrease in the measured distance}
    	\LineComment {\textit{thr} is set to a large number smaller than the vehicle's height }
        \If {($last_l$ - $dist_l$) > \textit{thr} \textbf{or} ($last_c$ - $dist_c$) > \textit{thr} \textbf{or} ($last_r$ - $dist_r$) > \textit{thr}}
            \State \Return \textbf{true}
        \Else
            \State $last_l$ $\gets$ $dist_l$, $last_c$ $\gets$ $dist_c$, $last_r$ $\gets$ $dist_r$
        \EndIf
	\EndWhile
\EndFunction
\end{algorithmic}
\end{algorithm}

Figure~\ref{fig3} shows examples of a laser triggering after sudden measurement changes.

\begin{figure}[H]
\includegraphics[width=\linewidth]{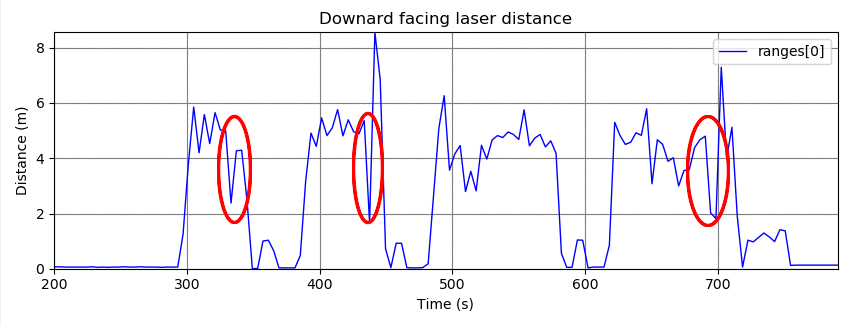}
\caption{Plot of laser distance measurements during a flight in our trials. The passing vehicle is detected in several moments, indicated by the red ellipses. Detection by this laser fails in the third trial due to vehicle misalignment with the laser direction.}
\label{fig3}
\end{figure}

\subsection{Landing Using Visual Servoing} \label{sec:landingvisualservo}

After the vehicle is detected (using the lasers or the visual pattern detection), the next task is trying to land on the defined pattern on the deck of the moving ground vehicle.
In our strategy, the UAV follows and lands the moving deck using visual servoing. Visual servoing consists of techniques that use information extracted from visual data to control the robot's motion. Assuming that the camera is attached to the drone through a gimbal, the vision configuration is called an eye-in-hand configuration. We used an Image-Based Visual Servoing scheme (IBVS), in which the error signal is estimated directly based on the 2D features of the target, which we considered to be the center of the target ellipse and the corners of its circumscribed rectangle. The error is then computed as the difference (in pixels) between the features' positions when the UAV is right above the target, at the height of about $50$~cm, and the current features' positions given by the deck detection algorithm (discussed in Section~\ref{sec:decktrack}). Since the robot operates in task space coordinates, there must be a mapping between the changes in image feature parameters and the robot's position. This mapping is applied using the image Jacobian, also known as the Interaction Matrix~\cite{espiau1992new}.

\subsubsection{Interaction Matrix}

If $\tau$ represents the task space and $F$ represents the feature parameter space, then the image Jacobian $L_v$ is the transformation from the tangent space of $\tau$ at $r$ to the tangent space of $F$ at $f$, where $r$ and $f$ are vectors in task space and feature parameter space, respectively. This relationship could be represented as: 

\begin{equation} \label{eq:vis-servo-relationship}
\dot{f} = L_v(r)\dot{r},
\end{equation}
\noindent which implies
\begin{equation}
L_v(r) = \left[
\frac{\partial f}{\partial r}
\right].
\end{equation}

In practice, $\dot{r}$ needs to be calculated given the value of $\dot{f}$ and the interaction matrix at $r$. Depending on the size and rank of the interaction matrix, different approaches can be used to calculate the inverse or pseudo-inverse of the interaction matrix, which gives the required $\dot{r}$ \cite{hutchinson1996visualservo}.

\subsubsection{Calculating Quadrotor's Velocity}

The developed visual servoing controller for this project assumes the presence of five 2D image features: the center of the detected ellipse $\mathbf{s}_c$ and the four corners of the rectangle that circumscribes the detected elliptic target. Since the gimbal has faster dynamics than the quadrotor, two separate controllers were developed. The first controller is a simple proportional controller for the gimbal pitch angle to keep the deck in the center of the image: 
\begin{equation}
\delta=k \left\|\mathbf{s}^\star_c-\mathbf{s}_c\right\|,
\end{equation}
\noindent where $k$ is a positive gain, $\mathbf{s}^\star_c$ is the desired location of the deck center in the image, and $\|\cdot\|$ stands for the Euclidean distance. Since the gimbal is not mounted at the center of the quadrotor, position $\mathbf{s}^\star_c$ has an offset to compensate for this.

The second control problem, which relies on image features $\mathbf{s}=[x, y]^T$, is a traditional visual servoing problem, which is solved using the following control law:
\begin{equation}
\mathbf{v}=-\lambda\left( \mathbf{L}_s {^c}\mathbf{V}_b {^b}\mathbf{J}_b \right)^+(\mathbf{s}^\star-\mathbf{s})+\mathbf{v}_{\rm ff}
\label{eqt:visualservoing}
\end{equation}
\noindent where $\mathbf{v}$ is the quadrotor velocity vector in the body control frame, $\lambda$ is a positive gain, $\mathbf{L}_s$ is the interaction matrix, ${^c}\mathbf{V}_b$ is a matrix that transforms velocities from the camera frame to the body frame, ${^b}\mathbf{J}_b$ is the robot Jacobian, $()^+$ is the pseudo-inverse operator, $(\mathbf{s}^\star-\mathbf{s})$ is the error computed in the feature space and $\mathbf{v}_{\rm ff}$ is a feed-forward term obtained by transforming the truck velocity to the body reference frame~\cite{marchand2005visp}. In our case, the error $(\mathbf{s}^\star-\mathbf{s})$ is minimized using only linear velocities and yaw rate. Therefore, $\mathbf{v}=[v_x, v_y, v_z, \omega]^T$, which forces the Jacobian matrix to be written as:
\begin{equation}
{^b}\mathbf{J}_b=
\begin{bmatrix}
1 & 0 & 0 & 0 \\
0 & 1 & 0 & 0\\
0 & 0 & 1 & 0\\
0 & 0 & 0 & 0\\
0 & 0 & 0 & 0\\
0 & 0 & 0 & 1\\
\end{bmatrix}.
\end{equation}

This matrix is responsible for the transformation from the general six-dimensional velocity space to a four-dimensional space that the UAV can indeed follow.

The interaction matrix $\mathbf{L}_s(\mathbf{s}^\star, Z^\star)$ in our method is constant, and is computed at the target location using image features ($\mathbf{s}^\star=[x^\star, y^\star]^T$) and the distance of the camera to the deck ($Z^\star$) as~\cite{espiau1992new}:
\begin{equation}
\mathbf{L}_s=
\setlength\arraycolsep{8pt}
\begin{bmatrix}
-1/Z^\star  & 0 & x^\star/Z^\star & x^\star y^\star & -(1+{x^\star}^2) & y^\star\\
0 & -1/Z^\star & y^\star/Z^\star & 1-{y^\star}^2 & -x^\star y^\star & -x^\star
\end{bmatrix}.
\end{equation}

It is essential to mention that, ideally, the interaction matrix should be computed online using the current features and height information. However, in our approach, we use a constant matrix to be more robust regarding the errors related to the UAV position estimation and information necessary to compute $Z$. For the same reason, the distance from the deck is not used as part of the feature, as is done in some visual servoing approaches~\cite{marchand2005visp}. In addition,  a constant matrix causes the system to be less sensitive to eventual spurious detection of the target. On the other hand, since the camera is mounted on a gimbal, matrix ${^c}\mathbf{V}_b$ must be computed online as: 
\begin{equation}  \label{eq:vis-servo-gimbal}
{^c}\mathbf{V}_b=
\begin{bmatrix}
{^c}\mathbf{R}_b & & [{^c}\mathbf{t}_b]_\times {^c}\mathbf{R}_b\\
\mathbf{0}_{3\times 3} & & {^c}\mathbf{R}_b
\end{bmatrix}
\end{equation}
\noindent where ${^c}\mathbf{R}_b$ is the rotation matrix between the image and the robot body, which is a function of the gimbal angles, ${^c}\mathbf{t}_b$ is the corresponding constant translation vector, and $[\mathbf{t}]_\times$ is the skew-symmetric matrix related to $\mathbf{t}$.

Since we are dealing with a moving deck, the feed-forward term $\mathbf{v}_{\rm ff}$ in Equation~(\ref{eqt:visualservoing}) is necessary to reduce the error $(\mathbf{s}^\star-\mathbf{s})$ to zero~\cite{corke1996dynamic}. In our case, $\mathbf{v}_{\rm ff}$  originates from the ground vehicle velocity vector, which is a vector tangent to the moving vehicle's path. Thus, the precise estimation of this vector would require the localization of the vehicle with respect to the track, which is a challenging task. To make such an estimation more robust to sensor noise, we relaxed this problem to compute $\mathbf{v}_{\rm ff}$ only on the straight line segments of the track, where we assume the ground vehicle speed vector to be constant for several seconds. This restriction causes our UAV to land only on such segments.

All the visual servoing control laws in Equations~(\ref{eq:vis-servo-relationship})--(\ref{eq:vis-servo-gimbal}) are implemented using the Visual Servoing Platform (ViSP)~\cite{marchand2005visp} version 3.0.1, which provides a suitable data structure for the problem and efficiently computes the required pseudo-inverse matrix.

\subsection{Initial Quadrotor Acceleration} \label{sec:initialaccel}

As described in Section~\ref{sec:landingstrategy}, the quadrotor will hover until it senses the ground vehicle passing below it. Due to the relatively slow dynamics of the quadrotor controller, it takes some time to accelerate to the ground vehicle's speed. By this time, the deck is already out of the camera's sight, and the visual servoing procedure cannot be used for landing.

To compensate for the delay in reaching the ground vehicle's speed and to gain sight of the deck again, an initial period of high acceleration is set right after sensing the passing ground vehicle. In this period, the desired speed is set to a value higher than the vehicle's speed until the quadrotor compensates for the created gap with the vehicle. After this time, the quadrotor switches to the visual servoing procedure described in Section~\ref{sec:landingvisualservo} with a feed-forward velocity lowered to the current ground vehicle's speed.

We tested two different criteria for switching from the higher-speed flight to the visual servoing (box 5 in Figure~\ref{fig2}) procedure: vision-based and timing-based. In the vision-based approach, the switch to visual servoing happens when the pattern on the moving vehicle is detected again. In the timing-based approach, depending on the processing and dynamics delays of the sensing and considering the known speed of the vehicle and the set speed for the UAV, it is possible to calculate a fixed time $t_a$ that is enough for catching up with the vehicle. After this time, the UAV can start the visual servoing landing.

\subsection{Final Steps and Landing}

When the UAV gets too close to the ground vehicle, the camera can no longer see the pattern. To prevent the UAV from stopping when the target detector (described in Subsection~\ref{sec:decktrack}) is unable to detect the deck during landing, the UAV is still commanded to move with the known vehicle's speed for a few sampling periods. If using one of its point lasers, the UAV detects that it has the correct height to land, it increases its downward vertical velocity and activates the drone landing procedure, which shuts down the propellers.

Note that the ``blind'' approach can only work due to the low delays between losing the pattern and landing (generally 30--70~ms). This approach may fail if the ground vehicle aggressively changes its direction or velocity. However, for many applications, the vehicle's motion (e.g., a ship or a delivery truck) can be reasonably assumed stable for such a short~time.

\subsection{Detection and Tracking of the Deck} \label{sec:decktrack}
Figure~\ref{fig1} shows the target pattern on top of the deck. There are some challenges in the detection of this pattern from the camera frames, including:

\begin{itemize}
\item This algorithm is used for landing the quadrotor on a moving vehicle. Therefore, it should work online (with a frequency greater than 10 Hz) on a resource-limited onboard computer.

\item The shape details cannot be seen in the video frames when the quadrotor is flying far from the deck.

\item The shape of the target is transformed by a projective distortion, which occurs when the shape is seen from different points of view.

\item There is a wide range of illumination conditions (e.g., cloudy, sunny, morning, evening). 

\item Due to the reflection of the light (e.g., from sources like the sun or bulbs), the target may not always be seen in all the frames, even when the camera is close to the vehicle.

\item In some frames, there may be shadows on the target shape (e.g., the shadow of the quadrotor or trees).

\item In some frames, only a part of the target shape may be seen.
\end{itemize}

Considering these challenges and the shape of the pattern, different participants of the MBZIRC challenge chose different approaches for detection. For example, in~\cite{bahnemann2018}, the authors implemented a quadrilateral detector for detecting the pattern from far distances and a cross detector for close-range situations. The method in~\cite{beul2018} uses the line and circular Hough-transform algorithms to calculate a confidence score for the pattern for the initial detection and then uses the rectangular area around the pattern for the tracking. In~\cite{jin2018}, a Convolutional Neural Network is developed to detect the elliptic pattern, which was trained with over 5000 images collected from the pattern moving at a maximum of $15$~km/h ($4.17$~m/s) at various heights. In~\cite{li2018} the cross and the circle are detected for far images, and only cross detection is used for the closer frames. The method uses the~\cite{fornaciari2014ellipsedetect} method for ellipse detection. In~\cite{tzoumanikas2018}, the outer square is detected first, and then the detection is verified by a template matching algorithm. 

In our work, we developed a novel method to overcome the problem challenges mentioned above, which detects and tracks the pattern by exploiting the structural properties of the shape without the need for any training~\cite{Keipour:2021:ral:ellipse}. More specifically, the deck detector system detects and tracks the circular shape (seen as an ellipse due to the projective transformation) on top of the deck, while ignoring the cross in the middle and then verifying that the detected ellipse belongs to the deck pattern. The developed real-time ellipse detection method can also detect the ellipses with partial occlusion or the ellipses that are exceeding the \mbox{image boundaries}. 

Figure~\ref{fig4} shows results for the detection of the deck in some sample frames. A more detailed description of the methods, database, and results are provided in \cite{Keipour:2021:ral:ellipse}.

\begin{figure}[H]\ContinuedFloat

\captionsetup[subfigure]{justification=centering}    
    \begin{subfigure}[b]{0.48\textwidth}
        \includegraphics[width=\textwidth]{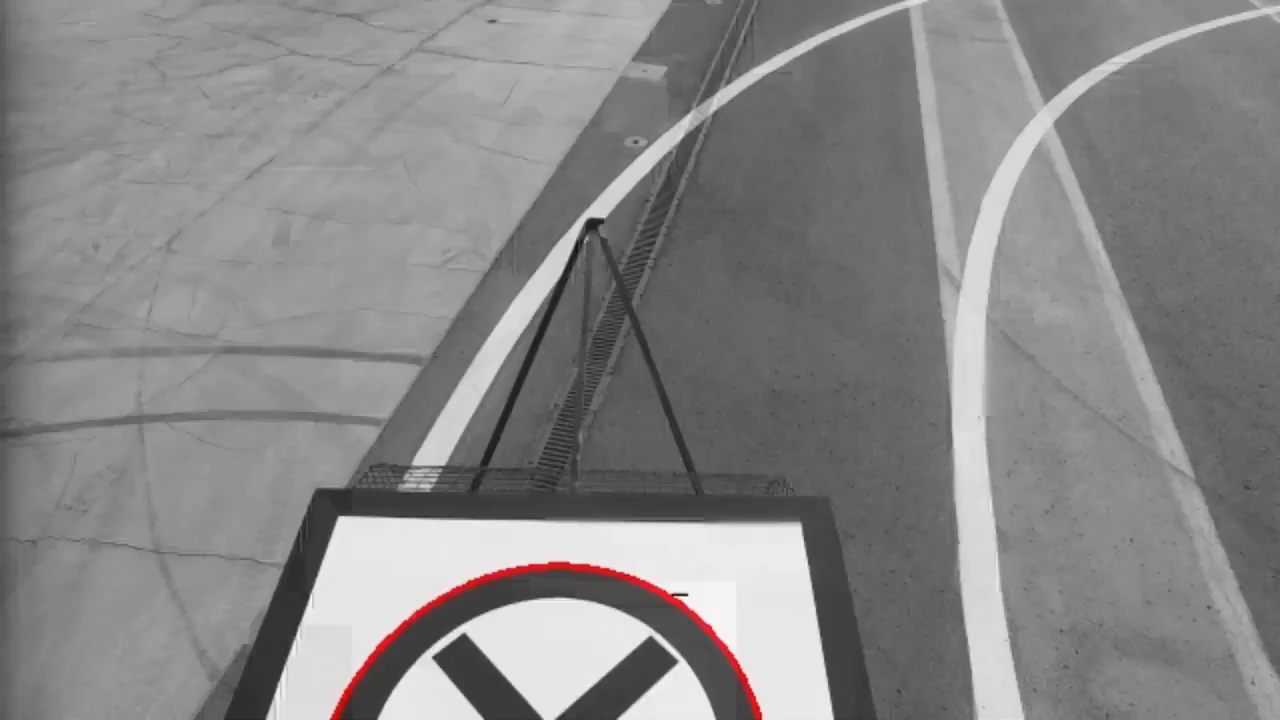}
        \caption{~}
    \end{subfigure}
    ~
    \begin{subfigure}[b]{0.48\textwidth}
        \includegraphics[width=\textwidth]{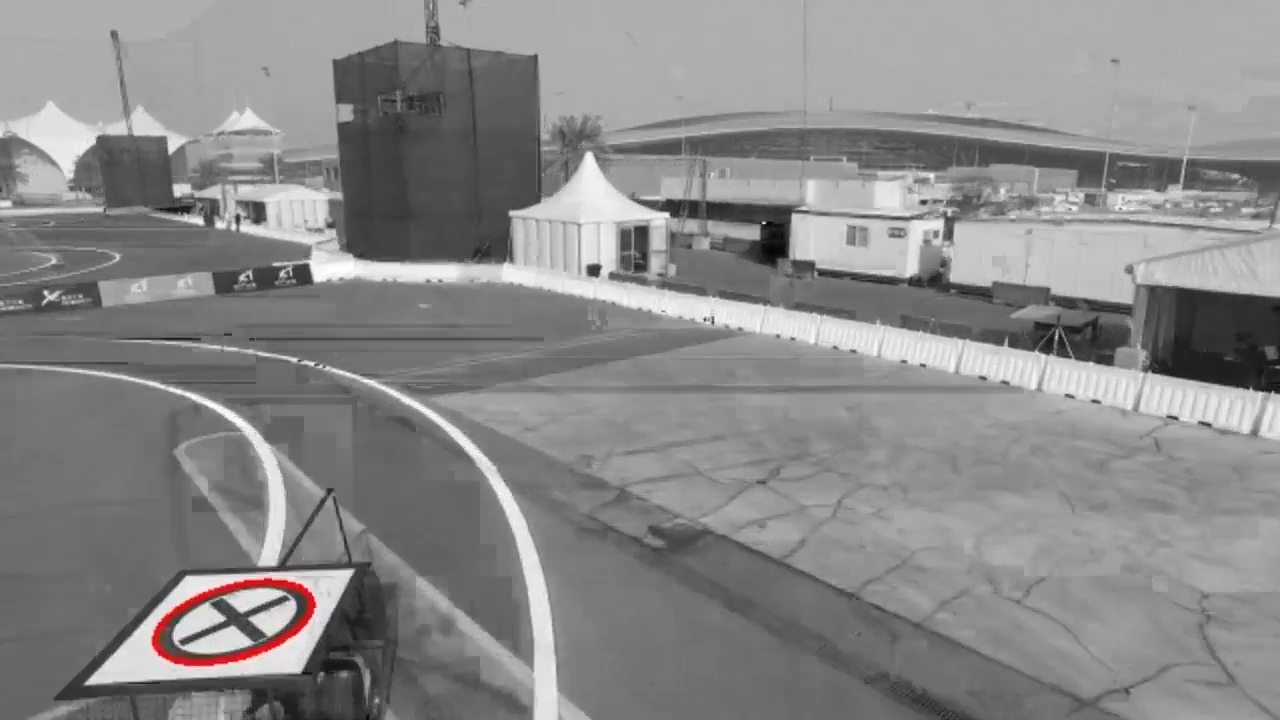}
        \caption{~}
    \end{subfigure}
    
    \vspace{6pt}
    
    \begin{subfigure}[b]{0.48\textwidth}
        \includegraphics[width=\textwidth]{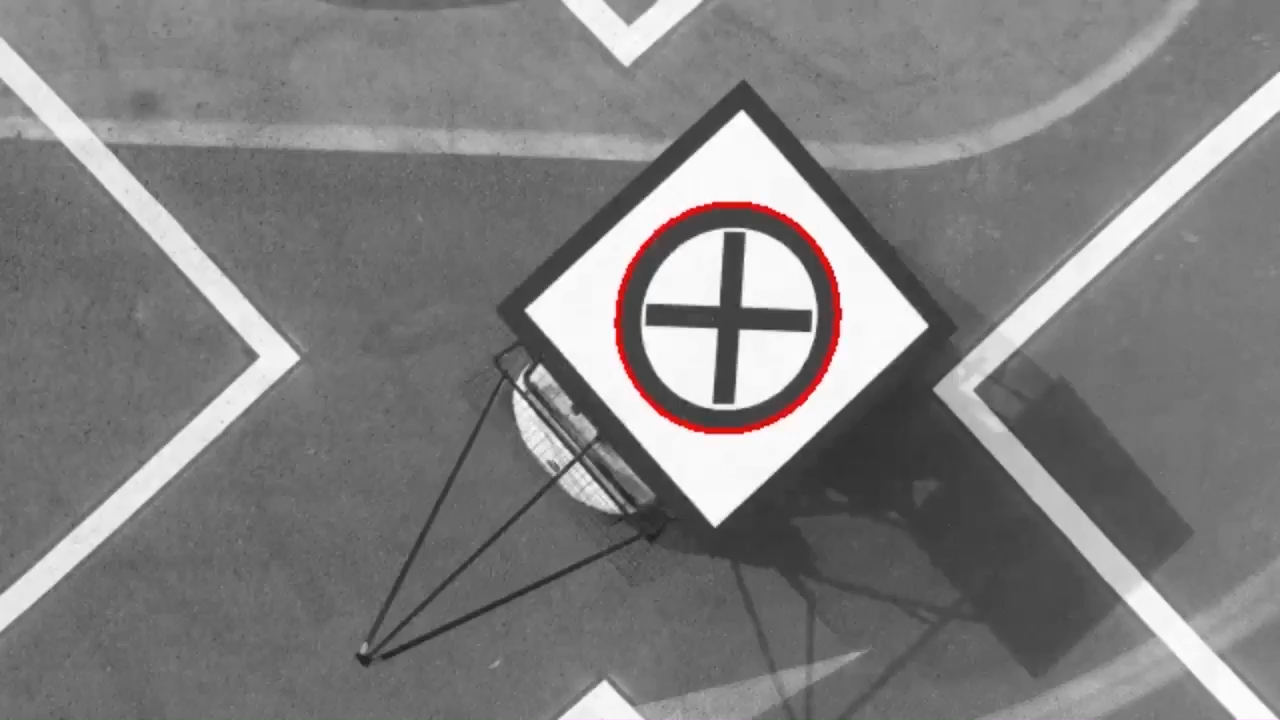}
        \caption{~}
    \end{subfigure}
    ~
    \begin{subfigure}[b]{0.48\textwidth}
        \includegraphics[width=\textwidth]{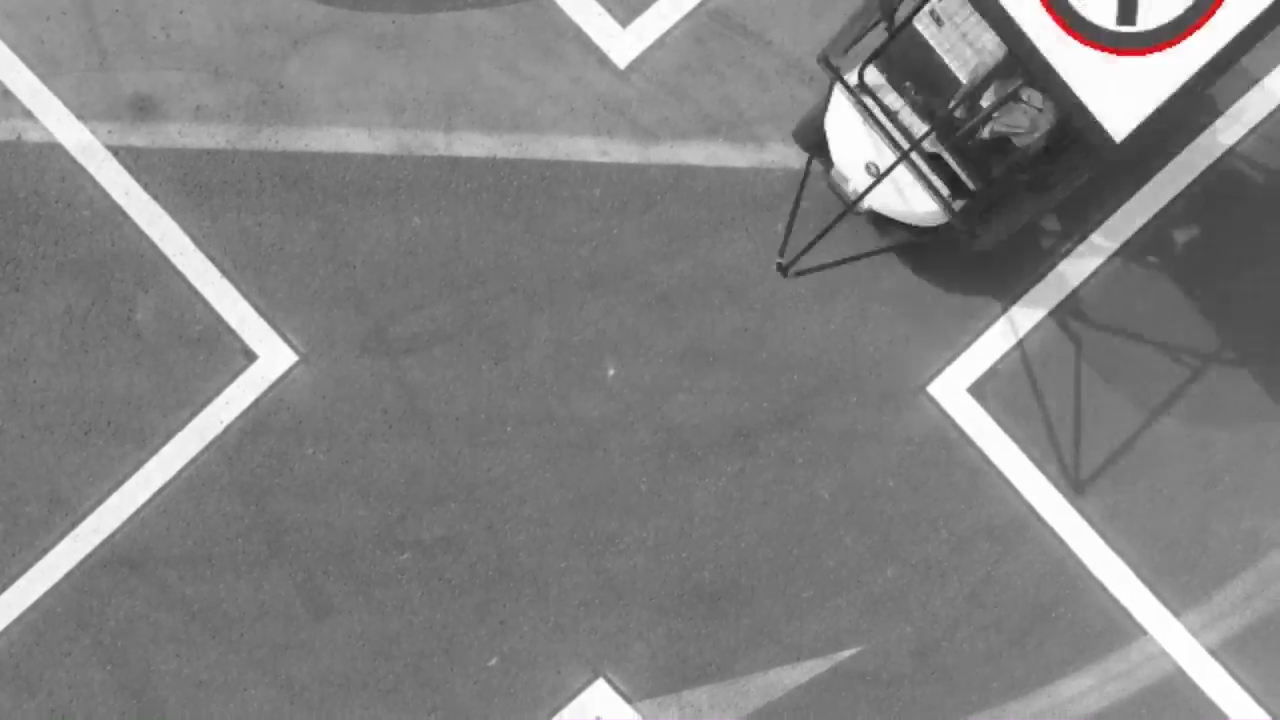}
        \caption{~}
    \end{subfigure}
\caption{Result of 
 the deck detection and tracking algorithm on sample frames. The red ellipse indicates the detected pattern on the deck of the moving vehicle.}
\label{fig4}
\end{figure}


\section{Experiments and Results} \label{sec:results}
This section outlines the architecture of our proposed landing method from both hardware and software perspectives and presents the results obtained from the tests in different conditions: simulation, indoor testing, and outdoor testing. Some videos of experiments and sequences shown in this section are available at 
 \url{http://theairlab.org/landing-on-vehicle}.

\subsection{Hardware} \label{sec:hardware}
The aerial platform chosen for this project is a DJI Matrice~100 quadrotor. 
This UAV is programmable, can carry additional sensors, is fast, provides velocity commands, and is accurate enough for this problem. Using an off-the-shelf vs. home-built platform vastly increased the speed of the development, reducing the time needed for dealing with hardware bugs. The selected platform has a GPS module for state estimation and is additionally equipped with an ARM-based DJI Manifold onboard computer (based on NVIDIA Jetson TK1 computer), a DJI Zenmuse X3 Gimbal and Camera system, a set of three SF30 altimeters, and four permanent magnets at the bottom of legs. One altimeter is pointed straight down to measure the current height of the flight with respect to the ground. The other two altimeters are mounted on the two sides of the quadrotor, pointing down with approximately $30$~degrees outward skew to measure the sudden changes in the height on the sides of the quadrotor for the detection of the truck passing next to the quadrotor. The camera is mounted on a three-axis gimbal and outputs grayscale images at a frequency of $30$~Hz with a resolution of $640\times360$~pixels.

\begin{figure}[H]
\includegraphics[width=0.8\linewidth]{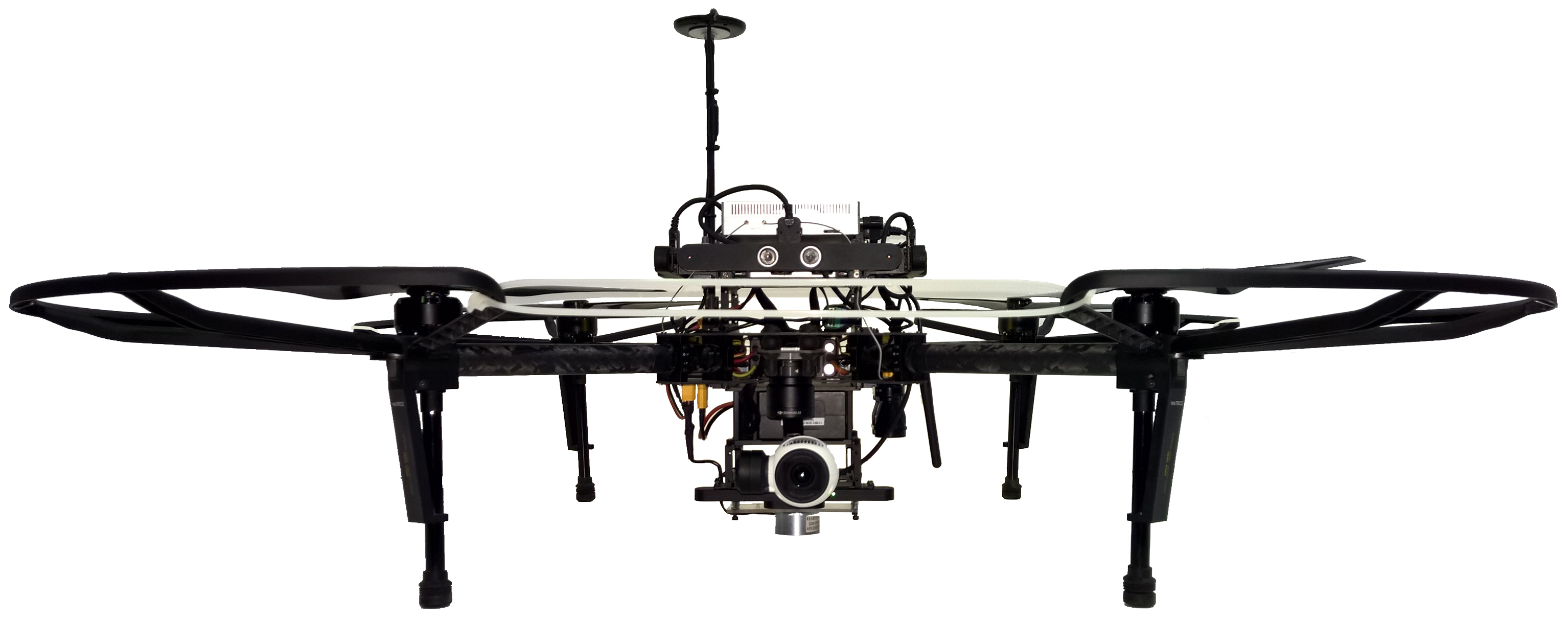}
\caption{The robot developed for our project. The figure shows our DJI Matrice 100 quadrotor equipped with the indoor testing parts (DJI Guidance System and the propeller guards), as well as the DJI Zenmuse X3 Gimbal and Camera system, the SF30 altimeter, and the DJI Manifold onboard~computer.}
\label{fig5}
\end{figure}

A wireless adapter allows communication between the robot computer and the ground station for development and monitoring. For indoor tests, a set of propeller guards and a DJI Guidance System, a vision-based system able to provide indoor localization, velocity estimation, and obstacle avoidance, are used to provide velocity estimation and improve the safety of the testing. The guards and the Guidance system are removed for outdoor tests to reduce the weight and increase the robustness to the wind. Figure~\ref{fig5} shows the picture of the robot.

The main characteristics of the robot are summarized in Table~\ref{tab1}.

\begin{table}[H]
\caption{The main 
 characteristics and parameters of the UAV.}
\label{tab1}
\newcolumntype{C}{>{\centering\arraybackslash}X}
\begin{tabularx}{\textwidth}{lC}
\toprule
\textbf{Parameter}   & \textbf{Value} \\ 
\midrule
Maximum time of flight & $19$~min \\ 
\midrule
Maximum horizontal speed  & $17$~m/s ($61$~km/h) \\ 
\midrule
Maximum vertical speed & $4$~m/s ($14.4$~km/h) \\ 
\midrule
External diameter (with propeller guards) & $1.2$~m \\ 
\midrule
Height & $0.46$~m \\ 
\midrule
Processor & Quad-core ARM CORTEX-A15 \\ 
\midrule
RAM Memory & $2~$GB \\ 
\midrule
Maximum robot speed & $8.33$~m/s ($30$~km/h) \\ 
\bottomrule
\end{tabularx}
\end{table}

\subsection{Software} \label{sec:software}

\begin{figure}[H]
\includegraphics[width=0.7\linewidth]{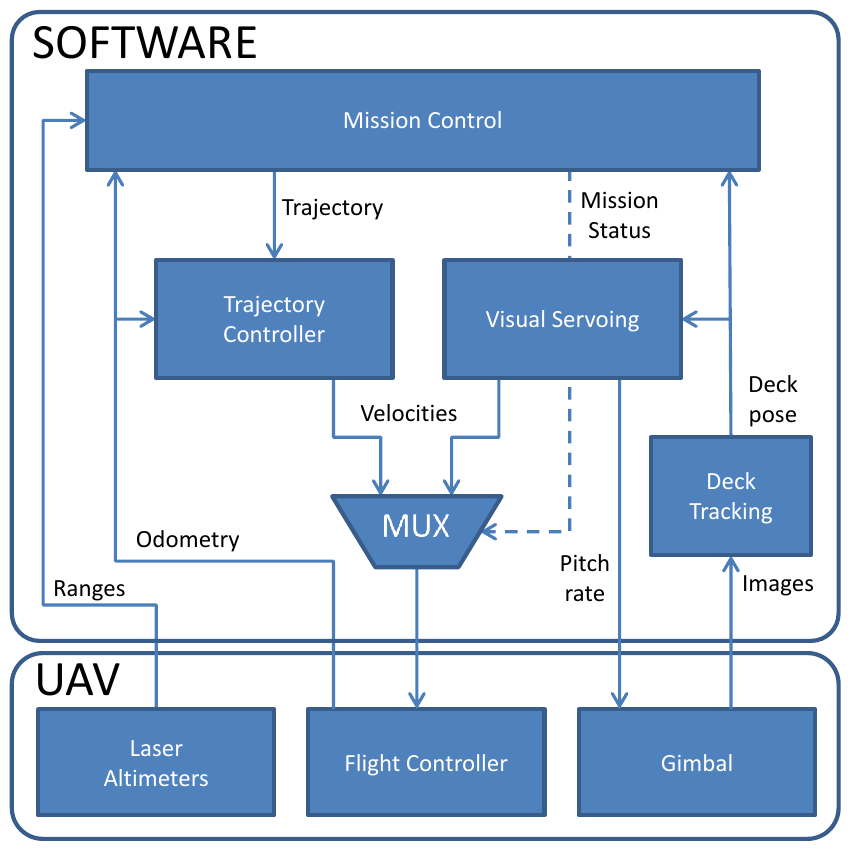}
\caption{System architecture used in the development of the system.}
\label{fig6}
\end{figure}

The robot's software is developed using the Robot Operating System (ROS) to achieve a modular structure. The modularity allowed the team members to work on different parts of the software independent of each other and reduced the debugging time. DJI's Onboard SDK is used to interact with the quadrotor's controller. The software is constructed in a way that it can control both the simulator and the actual robot with just a few modifications. Figure~\ref{fig6} shows a general view of the system architecture.

In Figure~\ref{fig6}, the main block of our architecture (implemented as a ROS node), called ``Mission Control'', dictates the robot's behavior, informing the other blocks of the current task (mission status) using a ROS Parameter. The Mission Control block also generates trajectories for the robot when the current mission mode requires the robot to fly to a different position. To dictate the robot's behavior, Mission Control relies on the robot's odometry information and the robot's distance to the ground or target. 

The other blocks in Figure~\ref{fig6} are:

\begin{itemize}

\item Deck tracking---detects the deck target and provides its position and orientation in the image reference frame;
\item Visual Servoing---controls the UAV to track and approach the deck; 
\item Trajectory controller---provides velocity commands to the robot so it can follow the trajectories generated by the Mission Control node;
\item Mux---selects the velocities to be sent to the quadrotor, depending on the \mbox{mission status}.
\end{itemize}

\textls[-15]{The implementation of the project and the datasets used for our tests are provided as open-source for public use on our website.} 

\subsection{Test Results} \label{sec:testresults}
Before testing the landing outdoors, we performed several indoor experiments to develop and validate our visual servoing approach.
We performed experiments both with a static and moving deck using this setup at speeds up to $2$~m/s. Figure~\ref{fig7} shows snapshots of one of our experiments. A video of the continuous sequence of experiments can be accessed from our website. 

\begin{figure}[H]
\captionsetup[subfigure]{justification=centering}
    \begin{subfigure}[b]{0.48\textwidth}
        \includegraphics[width=\textwidth]{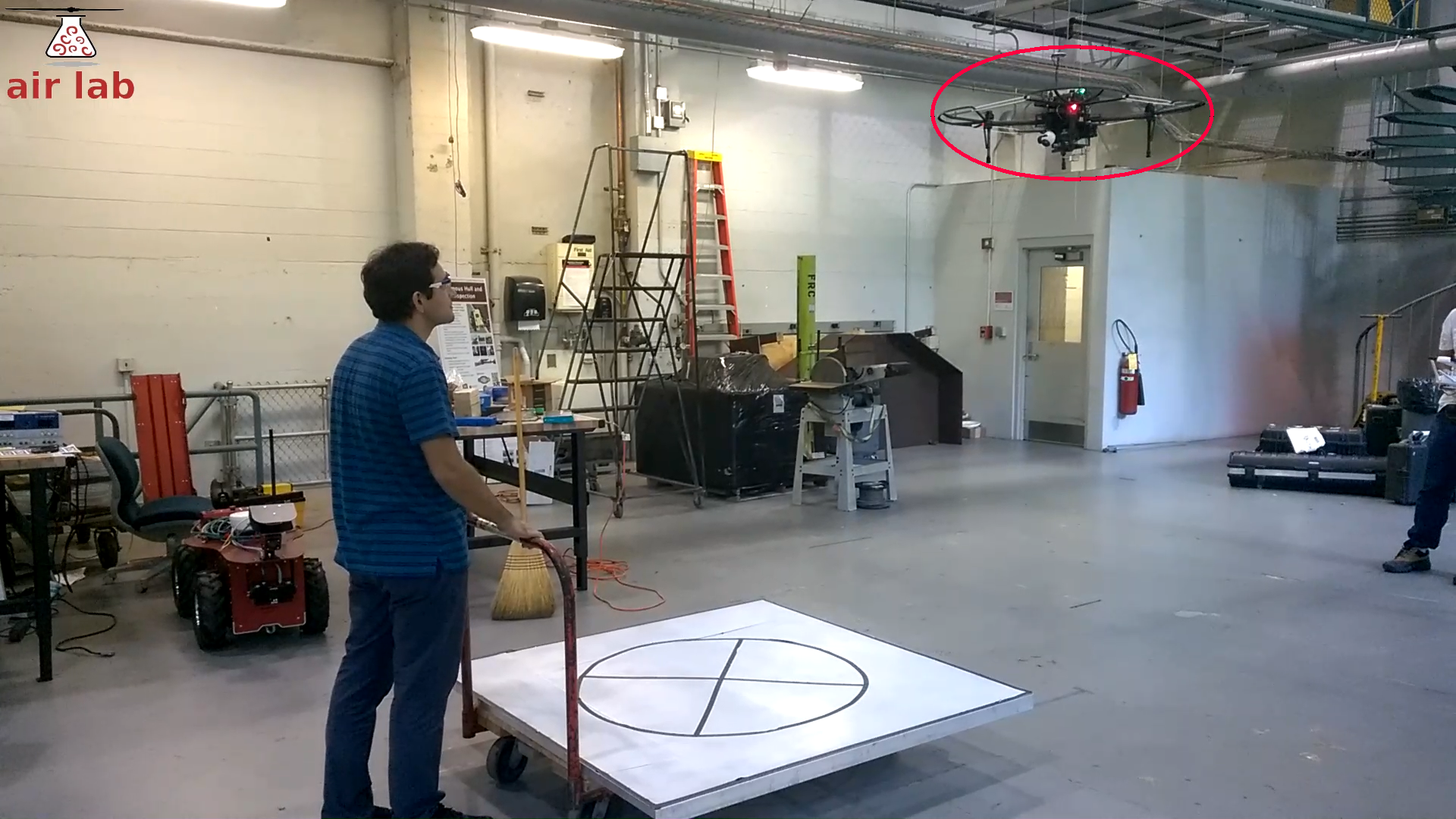}
        \caption{~}
   \end{subfigure}
    ~
    \begin{subfigure}[b]{0.48\textwidth}
        \includegraphics[width=\textwidth]{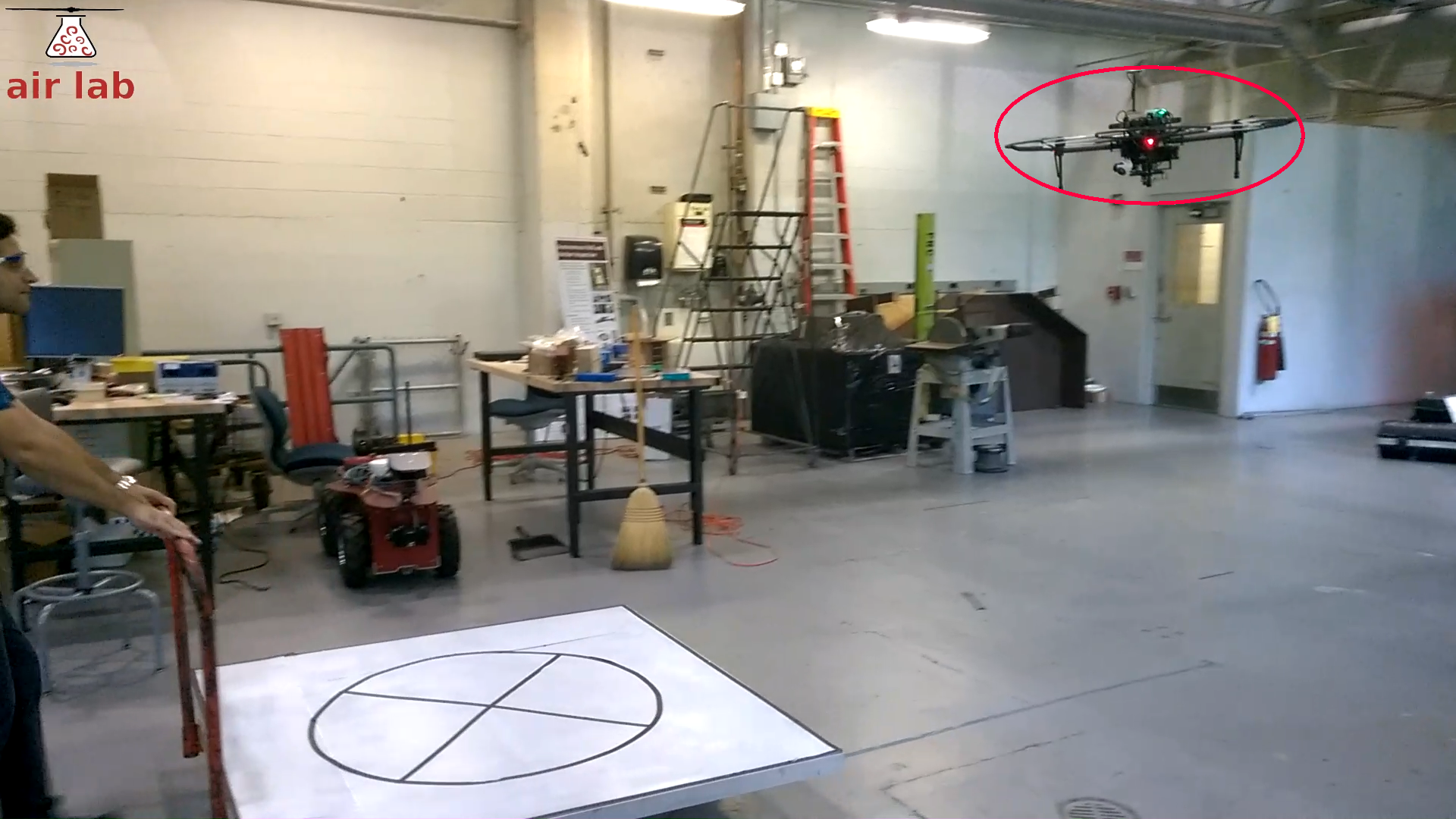}
        \caption{~}
    \end{subfigure}
    
    \vspace{6pt}
    
    \begin{subfigure}[b]{0.48\textwidth}
       \includegraphics[width=\textwidth]{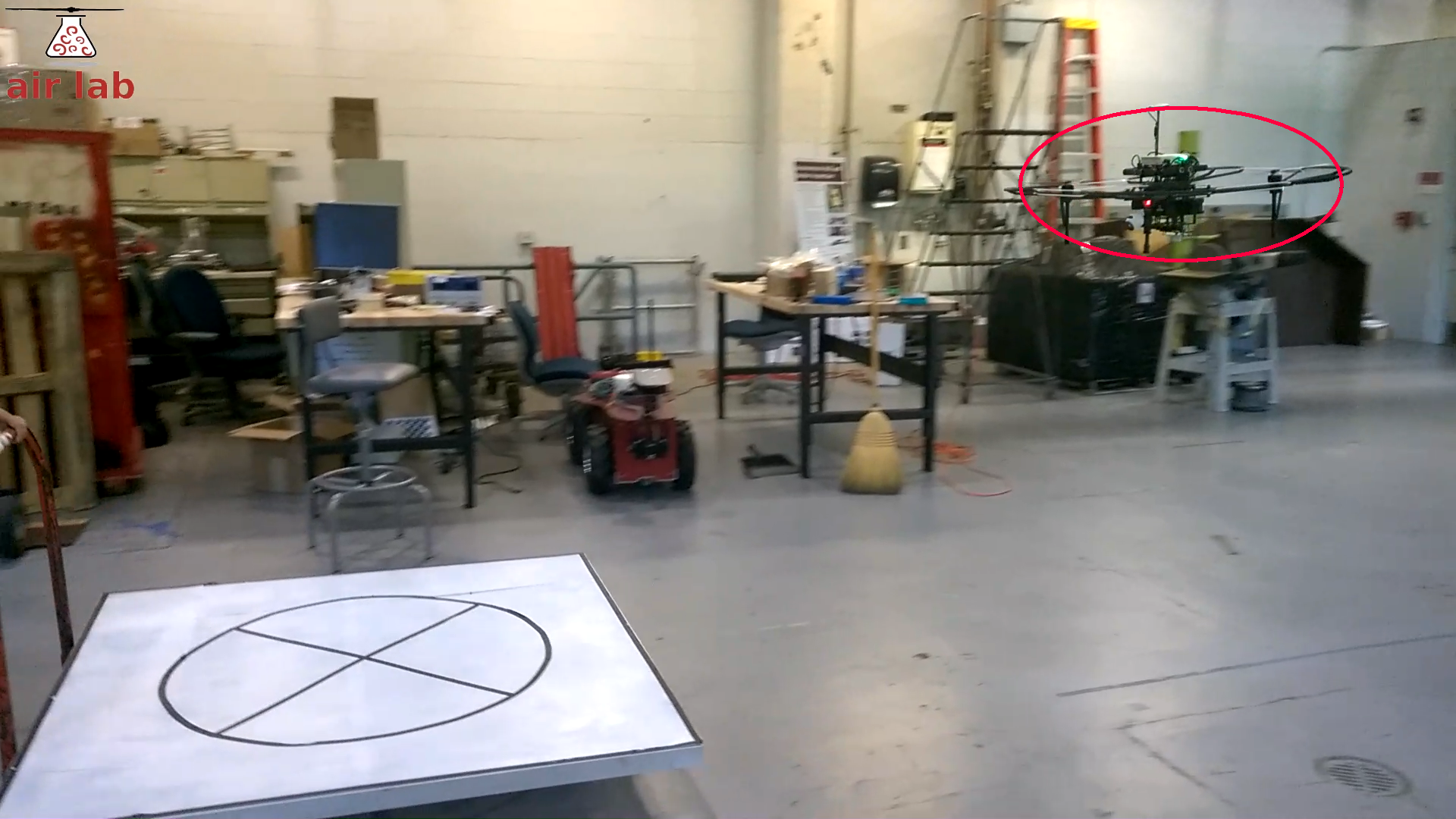}
        \caption{~}
    \end{subfigure}
    ~
    \begin{subfigure}[b]{0.48\textwidth}
        \includegraphics[width=\textwidth]{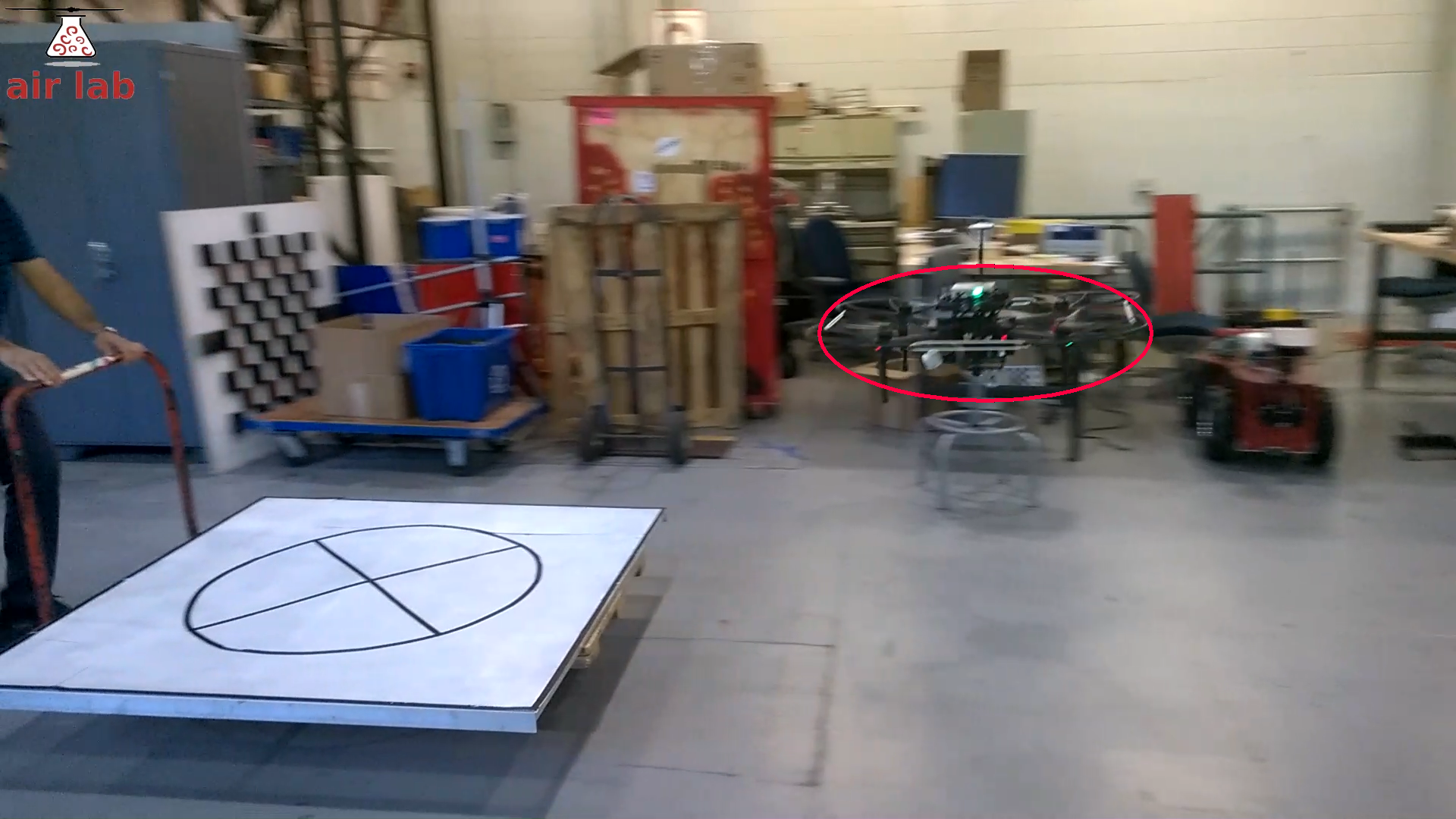}
        \caption{~}
    \end{subfigure}

    \caption{\textit{Cont}.}
\label{fig:petit3leaks3D1}
\end{figure}

\begin{figure}[H]\ContinuedFloat
\captionsetup[subfigure]{justification=centering}

    \begin{subfigure}[b]{0.48\textwidth}
        \includegraphics[width=\textwidth]{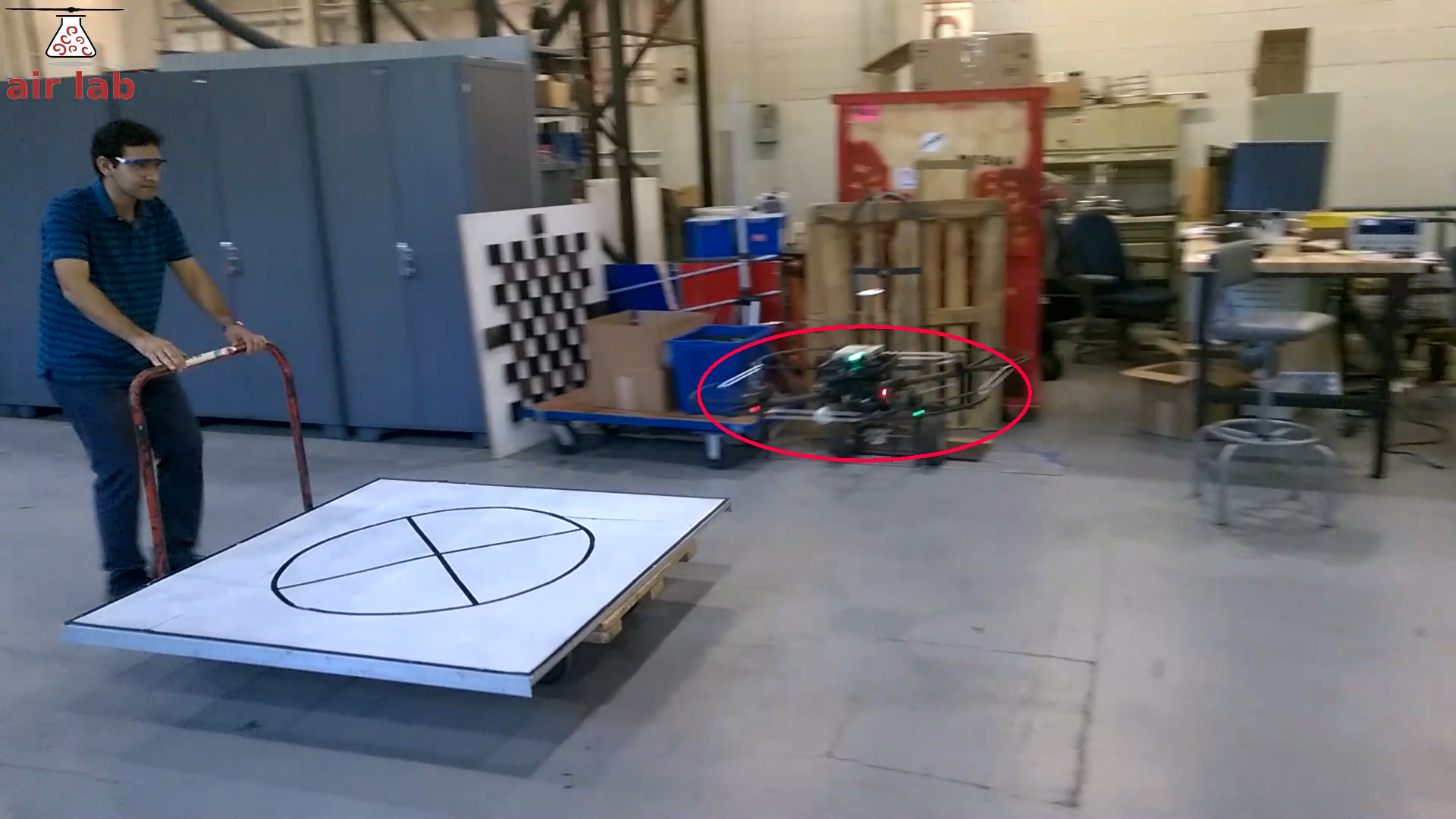}
        \caption{~}
    \end{subfigure}
    ~
    \begin{subfigure}[b]{0.48\textwidth}
        \includegraphics[width=\textwidth]{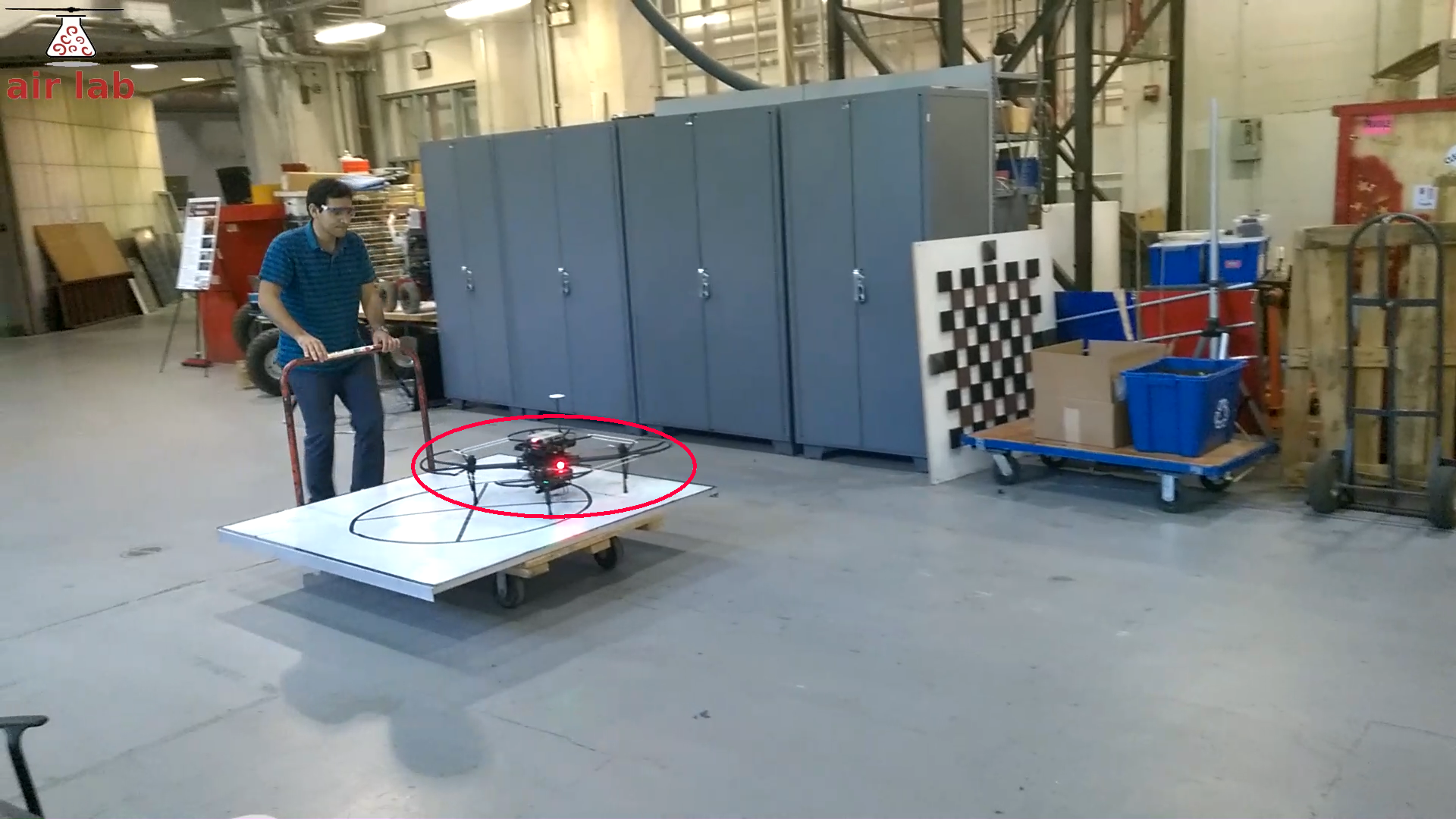}
        \caption{~}
    \end{subfigure}

\caption{Screenshots 
 from a video sequence showing our quadrotor landing on a moving platform. In this experiment, the quadrotor is additionally equipped with a DJI Guidance sensor for safe operation in the indoor environment.}
\label{fig7}
\end{figure}

Out of $19$ recorded trials to land indoors on a platform moving at a speed of 5--10~km/h (1.39--2.78~m/s), there were $17$ successful landings on the platform. One failure was due to loss of detection of the pattern, which happened midway to landing and the other failure was due to loss of the pattern at the last stage, which resulted in trying to land behind the~deck.

We developed a simulation environment using Gazebo \cite{koenig2004design}. It has a ground vehicle (truck) of $1.5$ m in height with the target pattern on top of it. The truck's motion is controlled by a ROS node that is parameterized by the truck speed and the direction of~movement.

Our UAV is simulated using the Hector Quadrotor~\cite{2012simpar_meyer}, which provides a velocity-based controller similar to the one provided by DJI's ROS SDK. The simulated quadrotor is also equipped with a (non-gimbaled) camera and a height sensor. Since the simulator has no gimbal, we fixed the camera on $45^\circ$. The odometry of the drone provides position estimates in the same message type provided by DJI. This similarity allows testing the entire software in simulation before actually flying the UAV. Therefore, the same software could be used both in simulation and on real hardware, provided that some parameters related to target detection were changed. Due to the kinematic nature of our controller, it is possible to run the trajectory controller with unchanged parameters.

\begin{figure}[H]
\includegraphics[width=0.7\linewidth]{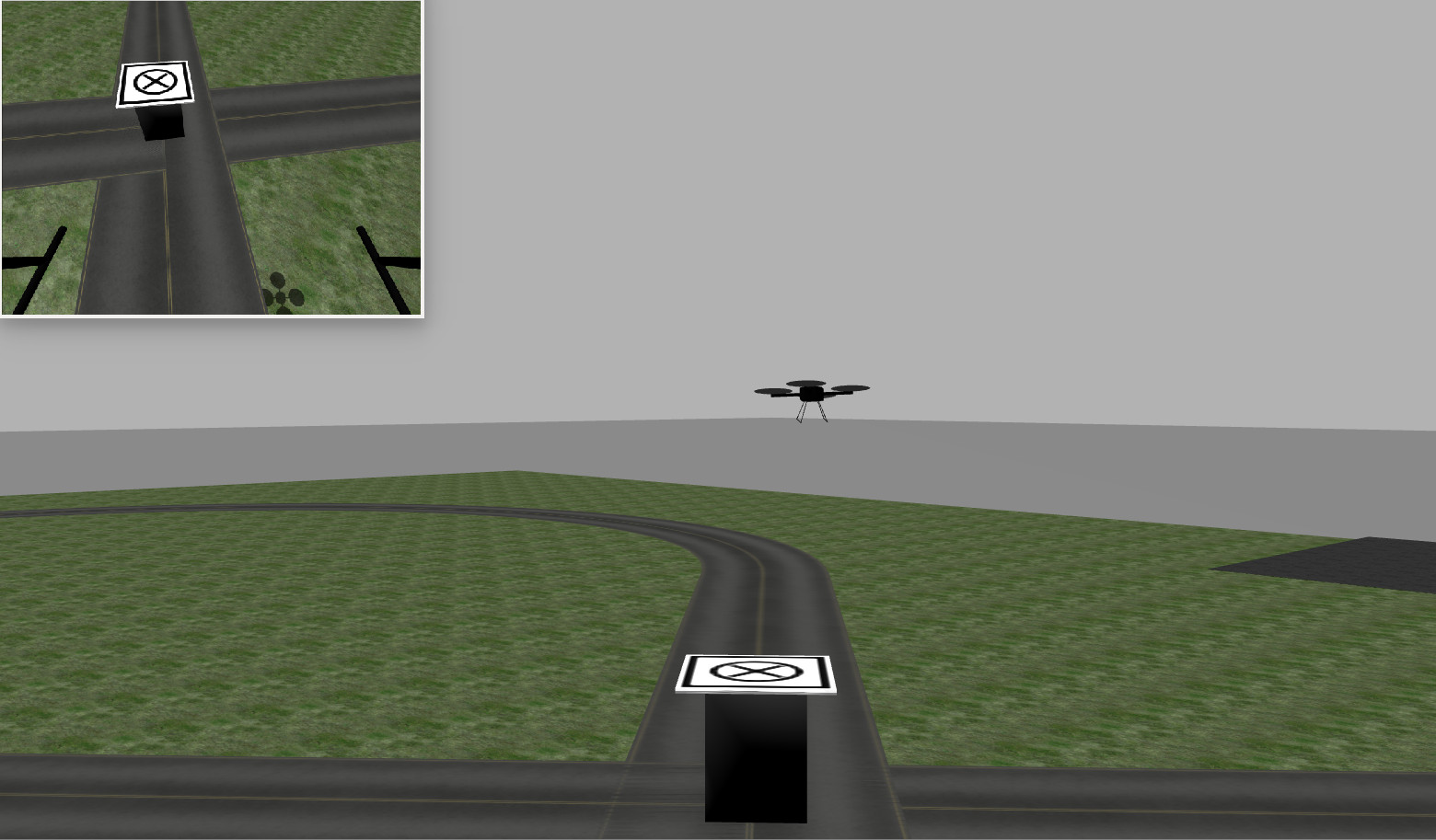}
\caption{A snapshot of the Gazebo simulation. The quadrotor's camera view is shown in the top left~corner.}
\label{fig8}
\end{figure}

Figure~\ref{fig8} shows a snapshot of the simulator when the quadrotor is about to land on the moving deck. The upper-left corner of the figure shows the UAV's view of the deck.

\begin{figure}[H]
\includegraphics[width=0.7\linewidth]{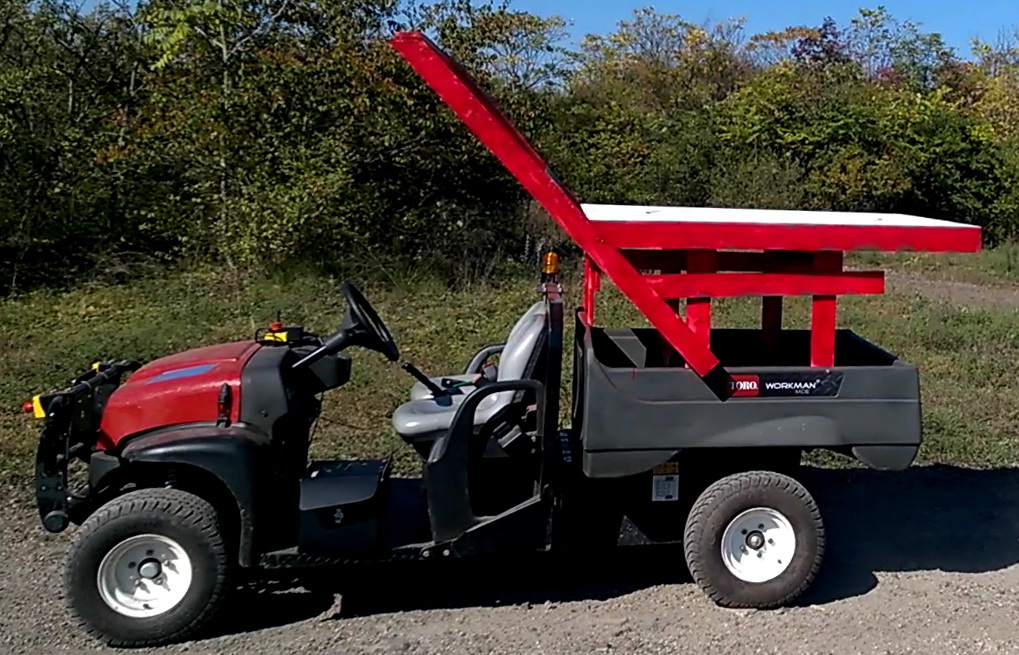}
\caption{The Toro Workman ground vehicle used for our tests.}
\label{fig9}
\end{figure}

To test the system in the real scenarios, a Toro MDE eWorkman electric vehicle was modified to support the pattern at the height of $1.5$ m (Figure~\ref{fig9}). A ferromagnetic deck with a painted landing pattern was attached to the top as the landing zone for the~quadrotor. 

The ground vehicle was manually driven with an approximate speed of $15$~km/h ($4.17$~m/s) measured by a GPS device. Visual servoing gains were empirically calibrated, and the robustness of the approach to landing was evaluated.

Although we did not collect detailed statistics on the landing procedure, there were several successful landings on the vehicle at the speed of $15$~km/h ($4.17$~m/s). It is hard to compare our method's success rate with the other methods, as none of the publications from the MBZIRC challenge have reported their success rates and have only reported limited results from the attempts during the challenge trials. 

Figure~\ref{fig10} shows a successful autonomous landing of the quadrotor on the moving vehicle used in our experiments at $15$~km/h ($4.17$~m/s) speed. 

Out of $22$ recorded trials, visual servoing was successful in bringing the UAV to the deck in $19$ cases. All three failure cases were due to the extreme sun reflection where the pattern was no longer seen in the frame; therefore, the visual servoing lost track of the pattern, and the mission was aborted. The cases where the switching from initial acceleration to visual servoing happened too soon or too late (which results in the loss of the target even at the beginning) were excluded from the analysis, as they were irrelevant to the study of the visual servoing performance. The results show the successful development of our visual servoing approach in bringing the UAV to the vicinity of the moving vehicle.

The time from the first detection of the platform to the stable landing varied between different runs from 5.8 to 6.5 s, which is less than the reported times of approaches claiming to be the fastest methods ($7.65$ s reported by \cite{tzoumanikas2018} and $7.8$ s reported by \cite{beul2018}). This timing can also be seen in the accompanying videos on our website. 

\begin{figure}[H]
\captionsetup[subfigure]{justification=centering}
    \begin{subfigure}[b]{0.48\textwidth}
        \includegraphics[width=\textwidth]{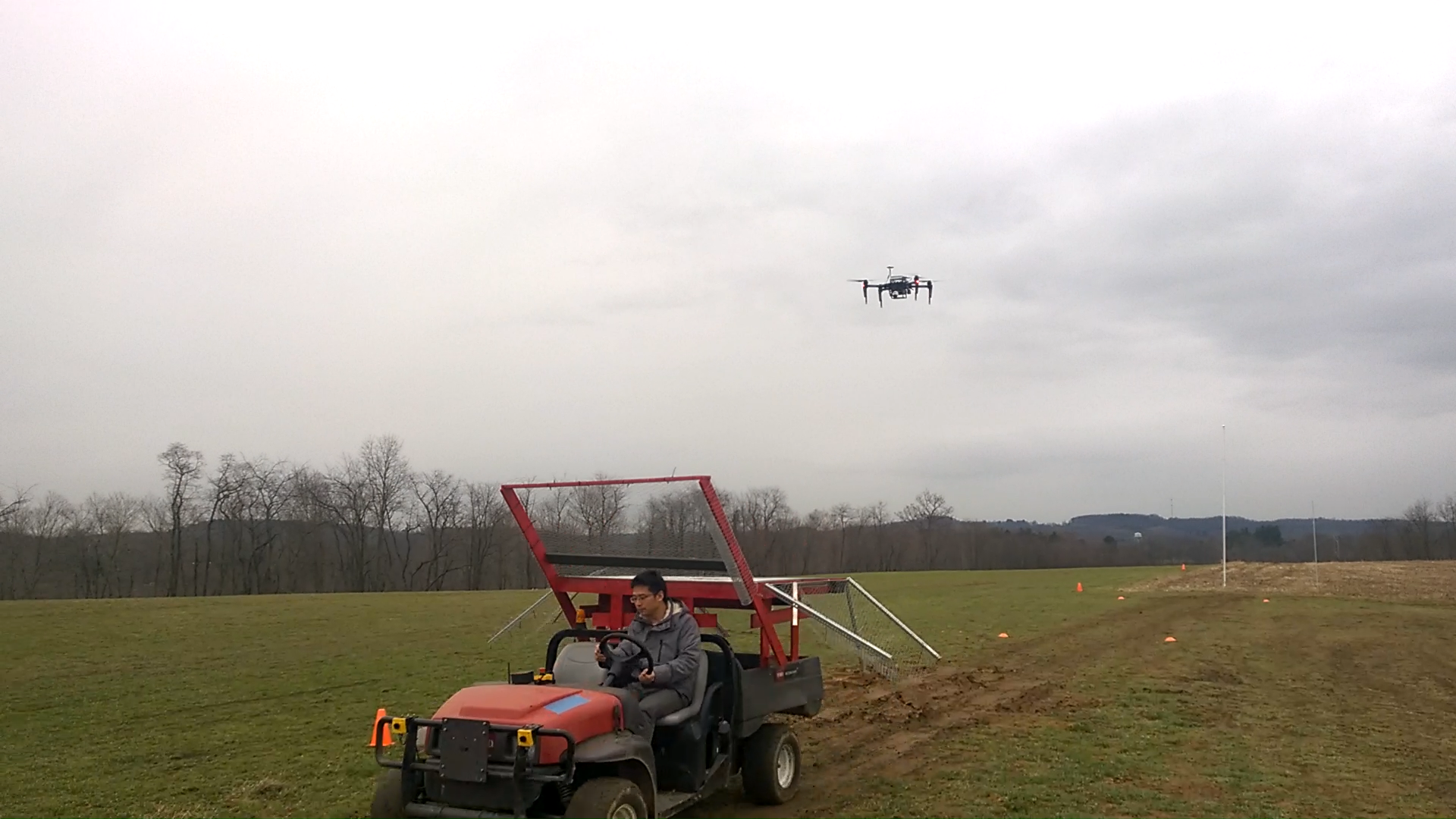}
        \caption{~}
   \end{subfigure}
    ~
    \begin{subfigure}[b]{0.48\textwidth}
        \includegraphics[width=\textwidth]{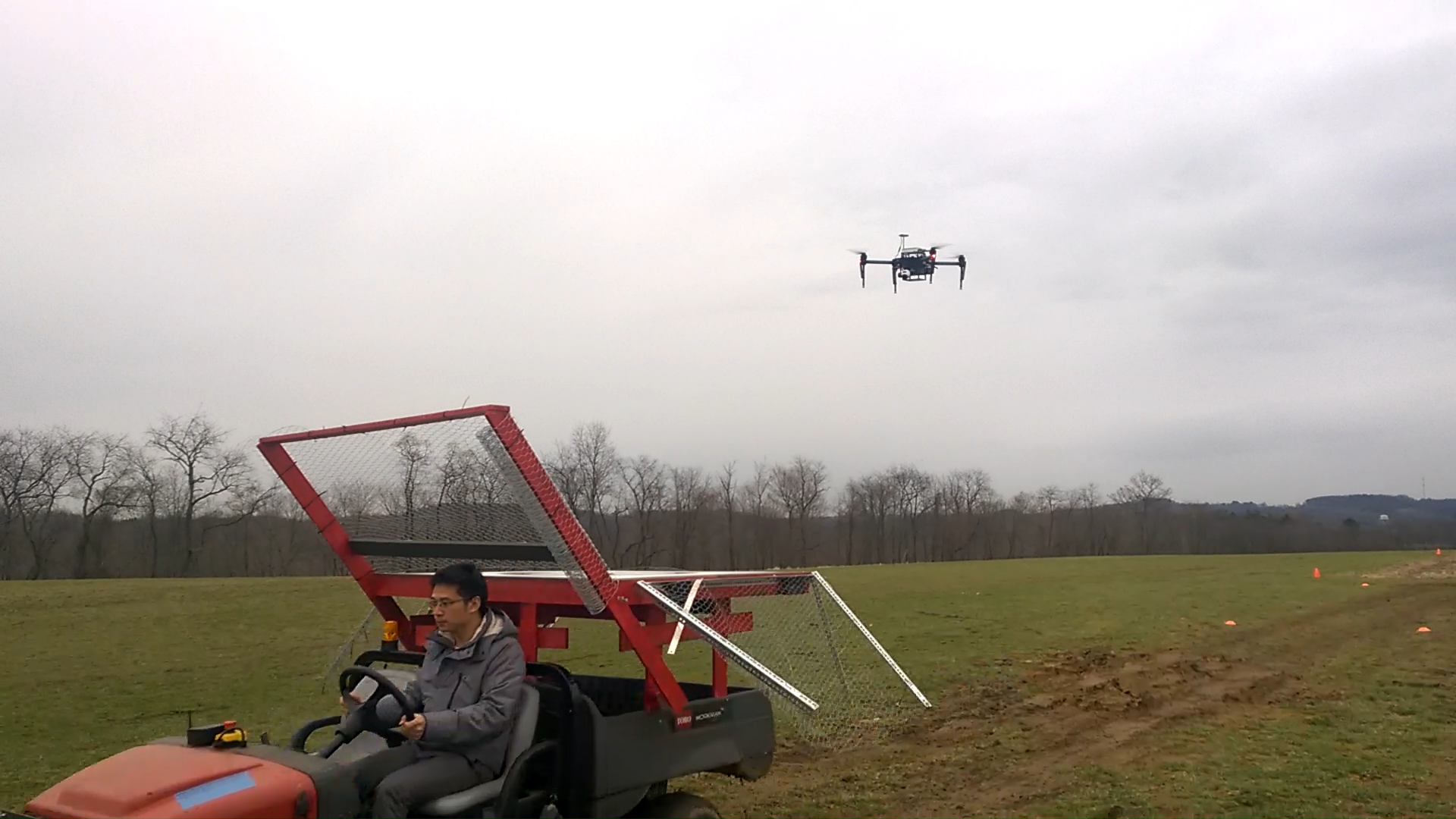}
        \caption{~}
    \end{subfigure}
    
    \caption{\textit{Cont}.}
\label{fig:petit3leaks3D2}
\end{figure}

\begin{figure}[H]\ContinuedFloat
 \captionsetup[subfigure]{justification=centering}
    
    \begin{subfigure}[b]{0.48\textwidth}
        \includegraphics[width=\textwidth]{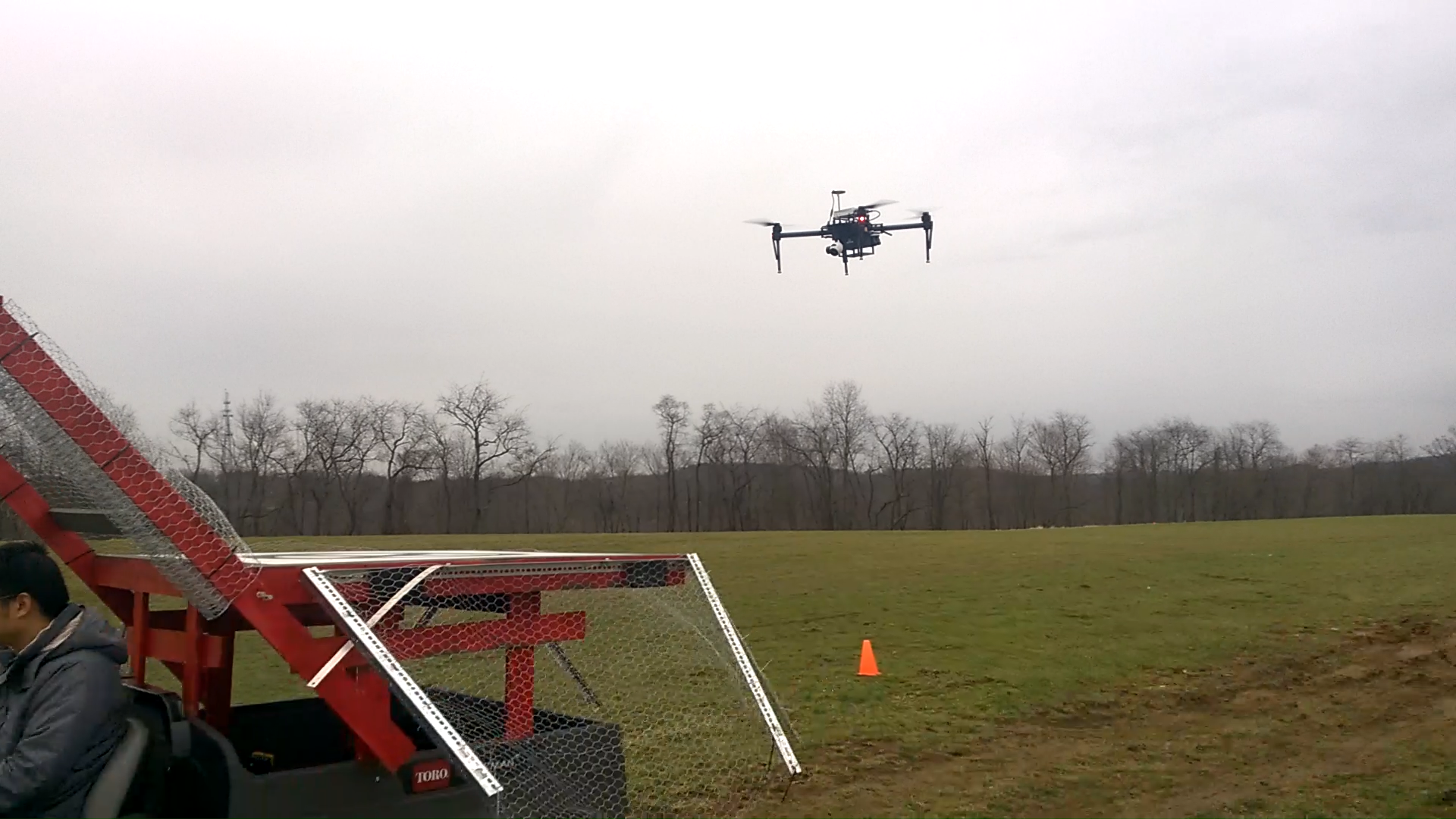}
        \caption{~}
    \end{subfigure}
    ~
    \begin{subfigure}[b]{0.48\textwidth}
        \includegraphics[width=\textwidth]{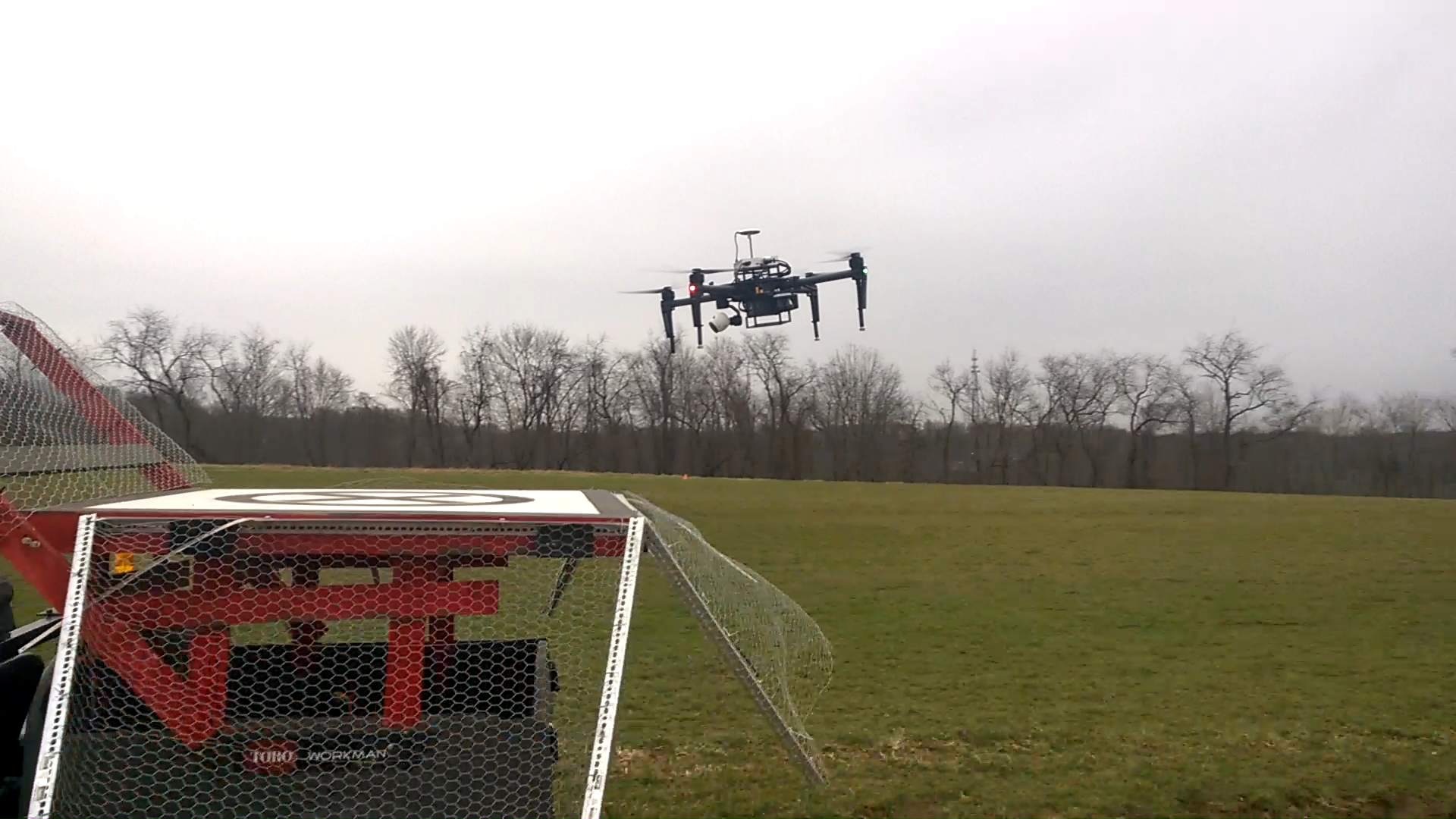}
        \caption{~}
    \end{subfigure}
    
    \vspace{6pt} 
    
   \begin{subfigure}[b]{0.48\textwidth}
        \includegraphics[width=\textwidth]{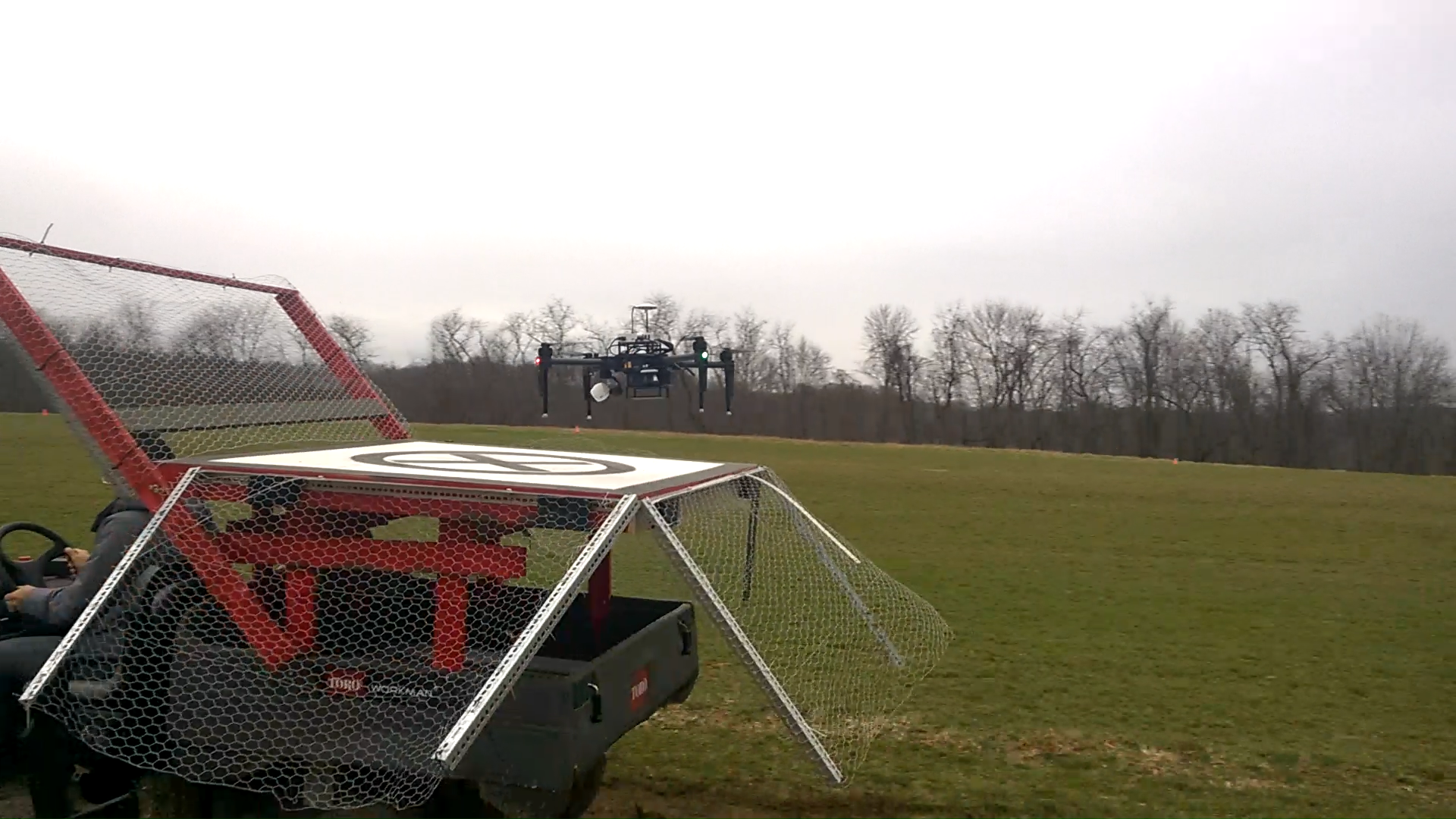}
        \caption{~}
    \end{subfigure}
    ~
    \begin{subfigure}[b]{0.48\textwidth}
        \includegraphics[width=\textwidth]{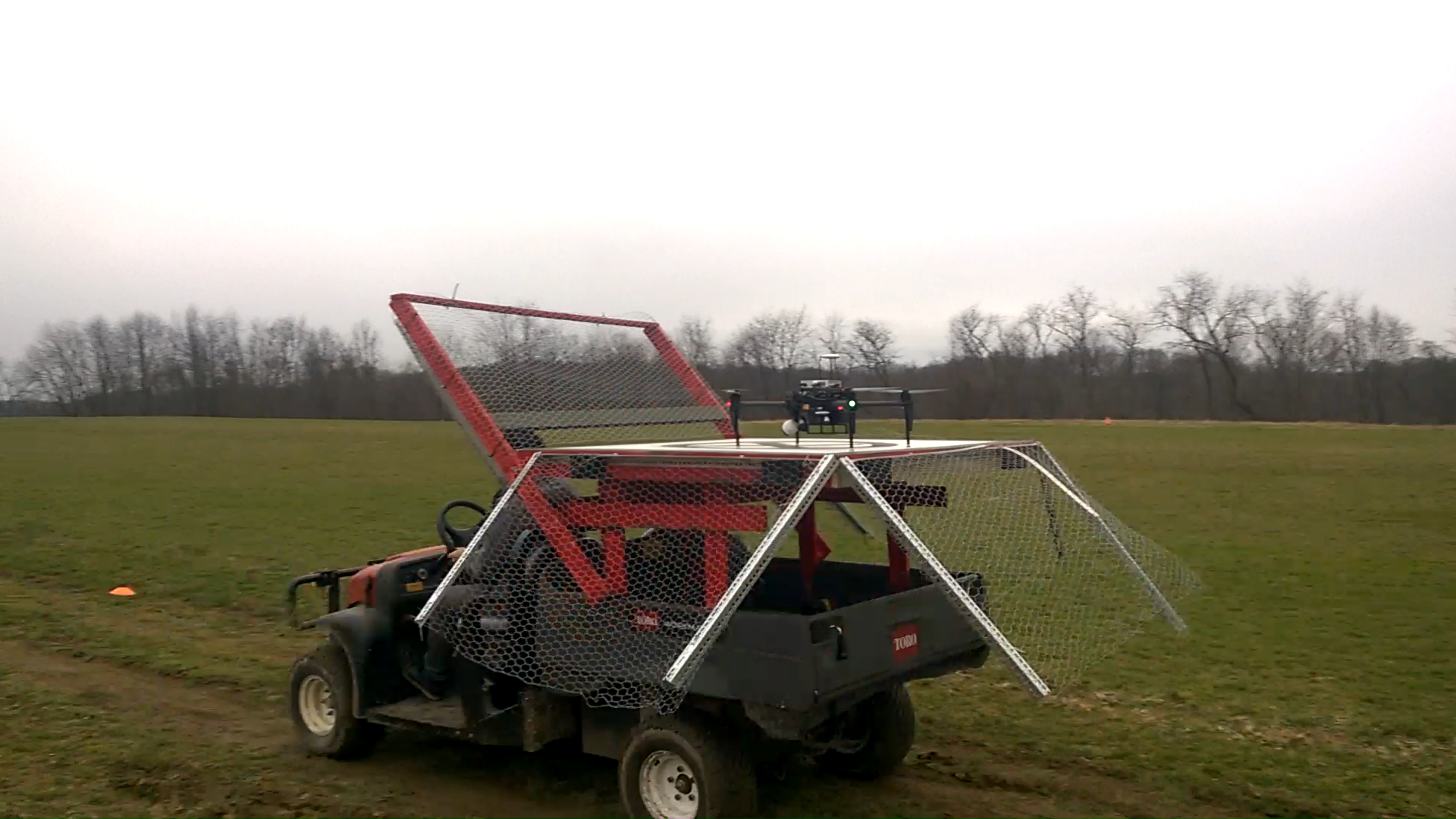}
        \caption{~}
    \end{subfigure}

\caption{Screenshots from 
a video sequence showing our quadrotor landing on the moving vehicle at $15$~km/h ($4.17$~m/s) speed. The video is available on our website.} 
\label{fig10}
\end{figure}


\section{Conclusions} \label{sec:conclusion}

This paper presented our approach to landing an autonomous UAV on a moving vehicle. With a few modifications, such as close-range landing zone detection and vehicle turn estimation, the proposed methods can work for a generalized autonomous landing scenario in a real-world application.

The visual servoing controller showed the ability to approach and land on the moving deck in simulation, indoor environments, and outdoor environments. The controller can work with a different set of localization sensors like GPS and DJI Guidance and does not rely on very precise sensors (e.g., motion capture system). 

The visual servoing controller showed promising results in reliably approaching the moving vehicle. A potential way to improve the landing efficiency further and tackle the loss of visual target problem is to divide the landing task into two subtasks: approaching the vehicle and landing from a short distance. After reaching the vehicle, the visual servoing approach can switch to a short-range landing algorithm. We believe that, by extending this approach, it can be used in a wide range of applications.

The robustness of the approach to occasional false positives in tracking the target and nondeterministic target detection rate has allowed us to use monocular vision with a real-time general-purpose ellipse detection method developed for this work (see~\cite{Keipour:2021:ral:ellipse}), further reducing the overall cost and weight of the system. However, as discussed in Section~\ref{sec:testresults}, this has caused the UAV to stop the landing procedure due to the sun reflection in 13.6\% of our outdoor tests. If this compromise cannot be accepted in an application, a more reliable target tracking method (such as infrared or radio markers) can be devised, or another attempt should be made at finding and approaching the vehicle.

While the approach has been tested for a range of vehicle velocities up to $15$~km/h, using the estimated vehicle speed as a feed-forwarding velocity means that the maximum ground vehicle's speed for UAV landing is mainly limited by the maximum speed of the UAV platform. However, the other potentially limiting factor can be the target detection rate and the gust factor, which at higher speeds can result in higher errors between two detections with insufficient time to correct the course.

\vspace{6pt} 



\clearpage
\authorcontributions{Conceptualization, A.K., G.A.S.P., and R.B.; methodology, A.K. and G.A.S.P.; software, A.K., G.A.S.P., R.B., R.G., and G.D.; validation, A.K., G.A.S.P., and S.S.; data curation, A.K., R.B., R.G., and G.D.; writing---original draft preparation, A.K., G.A.S.P., and R.B.; writing---review and editing, A.K., G.A.S.P., R.G., G.D., and S.S.; supervision, G.A.S.P. and S.S.; project administration, S.S.; 
 funding acquisition, S.S. All authors have read and agreed to the published version of the manuscript.}

\funding{The project was sponsored by Carnegie Mellon University Robotics Institute and Mohamed Bin Zayed International Robotics Challenge. During the realization of this work, Guilherme A.S. Pereira was supported by UFMG and CNPq/Brazil.}

\institutionalreview{Not applicable}

\informedconsent{Not applicable}

\dataavailability{The data presented in this study are openly available at 
 \url{http://theairlab.org/landing-on-vehicle}.} 

\acknowledgments{The authors want to thank Nikhil Baheti, Miaolei He, Zihan (Atlas) Yu, Koushil Sreenath, and Near Earth Autonomy for their support and help in this project. }

\conflictsofinterest{The authors declare no conflict of interest.} 



\abbreviations{Abbreviations}{
The following abbreviations are used in this manuscript:\\

\noindent 
\begin{tabular}{@{}ll}
UAV & Unmanned Aerial Vehicle\\
GPS & Global Positioning System\\
IMU & Inertial Measurement Unit\\
MBZIRC & Mohamed Bin Zayed International Robotics Challenge\\
RTK & Real-Time kinematic positioning\\
MPC & Model Predictive Control\\
RANSAC & Random Sample Consensus \\
RGB-D & Red, Green, Blue and Depth \\
MUX & Multiplexer \\
\end{tabular}
}



\begin{adjustwidth}{-\extralength}{0cm}

\reftitle{References}



\bibliography{references}
\end{adjustwidth}
\end{document}